\documentclass[11pt]{article}

\usepackage{amsmath}
\usepackage{amsmath}
\usepackage{amsfonts}

\usepackage[]{acl}

\usepackage{times}
\usepackage{latexsym}
\usepackage{algorithm, algpseudocode}
\usepackage[T1]{fontenc}

\usepackage[utf8]{inputenc}

\usepackage{microtype}
\usepackage{array}
\usepackage{booktabs}
\usepackage{graphicx}
\usepackage{subfigure}
\usepackage{longtable}
\usepackage{multicol}
\usepackage{geometry}
\usepackage{tabularx}
\usepackage{xtab,afterpage}
\usepackage{caption}
\usepackage{times}
\usepackage{latexsym}
\usepackage{amsmath,amssymb,amsfonts}
\usepackage{multirow}
\usepackage{hyperref}
\usepackage{dblfloatfix}
\usepackage{verbatim}
\usepackage{colortbl}
\usepackage{makecell}
\usepackage{longtable}
\usepackage{enumitem}

\newcolumntype{L}[1]{>{\raggedright\let\newline\\\arraybackslash\hspace{0pt}}m{#1}}
\newcolumntype{C}[1]{>{\centering\let\newline\\\arraybackslash\hspace{0pt}}m{#1}}
\newcolumntype{R}[1]{>{\raggedleft\let\newline\\\arraybackslash\hspace{0pt}}m{#1}}

\algnewcommand\algorithmicbreak{\textbf{break}}
\algnewcommand\Break{\algorithmicbreak}

\newcommand{\FrameworkName}{LoQ}
\newcommand{\FrameworkFull}{\textbf{L}ist \textbf{o}f \textbf{Q}uestions}
\newcommand{\Framework}{\textsc{\FrameworkName}}

\newcommand{\FeedQName}{FeedQ}

\newcommand{\FeedQ}{\textsc{\FeedQName}}

\newcommand{\QG}{\textsc{QG}}

\newcommand{\PredModel}{\textsc{PM}}

\usepackage{subfiles}
\usepackage{tcolorbox}

\usepackage{inconsolata}

\usepackage{graphicx}

\usepackage{listings}
\usepackage{xcolor}
\tcbuselibrary{breakable}

\newtcolorbox{promptbox}[1][]{
    colback=gray!5,
    colframe=teal!80,
    sharp corners,
    boxrule=0.8pt,
    breakable,
    #1
}

\newtcolorbox{examplebox1}[1][]{
    colback=gray!5,
    colframe=blue!30,
    sharp corners,
    boxrule=0.8pt,
    breakable,
    #1
}

\newtcolorbox{examplebox2}[1][]{
    colback=gray!5,
    colframe=red!30,
    sharp corners,
    boxrule=0.8pt,
    breakable,
    #1
}

\usepackage{xcolor}

\newcommand{\updt}[1]{\textcolor{black}{#1}}

\title{Improving Information Extraction with Learned Queries}

\author{Omar Sharif, Soroush Vosoughi, Nikhil Singh\\
Department of Computer Science\\
Dartmouth College \\
 \texttt{\{omar.sharif.gr, soroush.vosoughi, nikhil.u.singh\}@dartmouth.edu}}

\begin{document}
\maketitle
\begin{abstract}

   When information extraction fails, a natural instinct is to improve the model doing it: for example, by scaling it up or refining its reasoning. In this paper, we show that another part of the pipeline matters at least as much: the queries used to elicit this information. Across four clinical benchmarks and five LLMs, improving the question design alone raises performance by $\approx$18.6 F1-score points, i.e. more than using larger extraction models. To make such \textit{question design} learnable, we introduce \FrameworkFull~(\textbf{\Framework}), which generates document-specific question sets, and \FeedQ, a feedback-driven optimization method that iteratively refines questions against extraction outcomes. The resulting optimized questions can be used to train lightweight generators: with fine-tuning, 4B-parameter models match or outperform expert-derived baselines and substantially exceed the performance of much larger untuned models. We release a dataset of 12,820 optimized questions to support a broader shift in information extraction research toward treating question design as a first-class problem.
   
   
  

\end{abstract}
\section{Introduction}

\begin{quote}
    \textit{``If you do not know how to ask the right question, you discover nothing.''}\\
    \hfill --- W. Edwards Deming
\end{quote}

 We are surrounded by documents that contain answers to questions we have not yet asked. When those questions eventually arise, for example ``Which medication was prescribed in each of these health records?'' the task becomes one of \textit{information extraction}: identifying and structuring the relevant pieces of information embedded in the text. Straightforward as this goal may appear, it can be surprisingly difficult. The information of interest is rarely presented explicitly or in a consistent form; rather, it is often distributed across sentences, expressed indirectly, or intertwined with narrative context. Thus, to get the \textit{right} answers, we must ask the right questions.

This challenge is especially acute in clinical and biomedical text. Case reports, pharmacovigilance records, and discharge summaries convey structured facts through unstructured narratives \cite{pmlr-v287-zhang25b,10.1093/jamiaopen/ooaf109}. Consider extracting just one type of information: \textit{``Adverse Drug Reaction"}. Symptoms may be mentioned in one paragraph, a cause implied through temporal proximity to a medication, and a resolution described later still. Extracting such information reliably depends on both whether a model is capable and whether it is queried in a way that exposes the relevant evidence.

\begin{figure}[t]
  \centering
 \includegraphics[width=\linewidth]{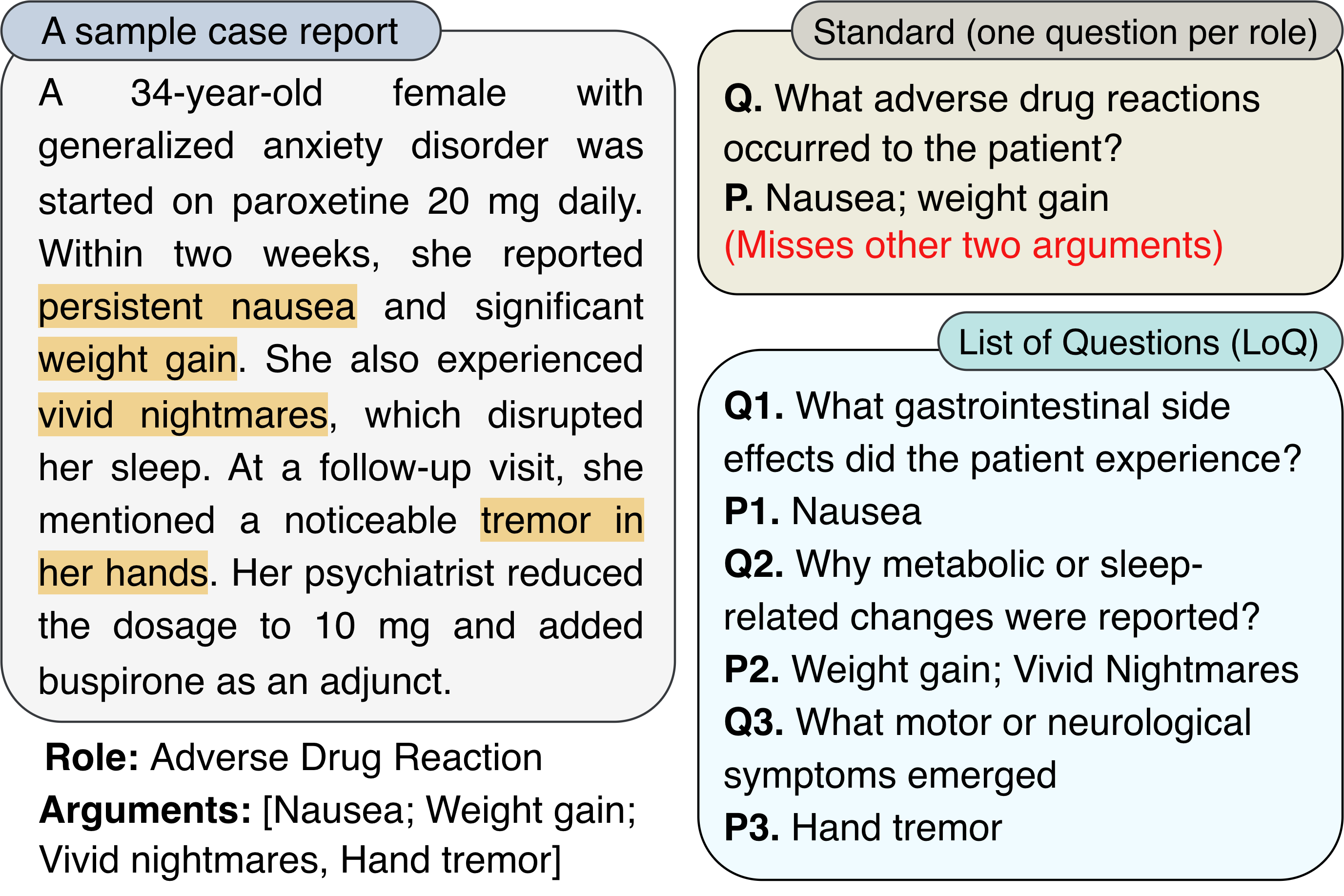}
 \caption{A case report describes a patient who develops four adverse drug reactions after starting paroxetine (highlighted). The standard approach asks a single role-level question and retrieves only two. \Framework~generates a set of document-grounded questions, each targeting a different symptom category, and recovers all four. The extraction model is identical in both settings; only the questions change, showing how extracting answers benefits from formulating the right questions. Additional examples are given in Table \ref{tab:question-examples}.}
 \label{fig:example}
\end{figure}

Large language models (LLMs) have become the dominant method for such tasks \cite{shimizu-etal-2025-exploring,tanev-etal-2025-exploring}. For LLMs this problem is often framed as question-based argument extraction: given a document and question about target role, the model generates the relevant arguments as free text \cite{sharif-etal-2024-explicit,zhang-etal-2024-ultra}. Here, a role is the \textit{type} of information to be extracted (e.g., \textit{Adverse Drug Reaction}), and arguments are the specific details for that role (e.g., the drug involved, the symptoms observed, the outcome). This formulation is flexible but it quietly inherits an assumption that the questions themselves are adequate. The dominant approaches derive questions from annotation schema or role-level templates and apply them uniformly across documents regardless of how the target information is expressed \cite{du-cardie-2020-event,sharif-etal-2025-regen}. As shown in Figure \ref{fig:example}, asking \textit{``What adverse drug reaction occurred to the patient?''} treats the role as a single retrieval target, assuming all relevant arguments can be surfaced with one generic query. This risks systematically missing arguments that require targeted attention. The bottleneck, then, is not only the LLMs' ability to answer but the way we ask question and direct this attention.

We introduce \FrameworkFull~(\textbf{\Framework}), a \textbf{framework for learning extraction questions tailored to each document--role pair}. \Framework~uses \FeedQ, an iterative loop that refines candidate questions based on the quality of the extractions they elicit. Optimized questions then help train a lightweight question generation model for deployment. Our contributions are thus:
\begin{itemize}

 \item We provide \textbf{empirical evidence} that question design is a major bottleneck in LLM-based information extraction. Across four clinical benchmarks (CaseReportBench \cite{pmlr-v287-zhang25b}, PHEE \cite{sun-etal-2022-phee}, DiscourseEE \cite{sharif-etal-2024-explicit}, and MACCROBAT \cite{ma-etal-2023-dice}) and five LLMs, improving questions can increase F1 by \textbf{18.6}\% on average.
 

\item We introduce a \textbf{two-part framework}, \Framework, for learning to generate effective, document-grounded extraction questions:
\begin{enumerate}
    \item First, we develop \FeedQ, a \textbf{feedback-driven iterative refinement method} for optimizing extraction questions without annotated question supervision. We use \FeedQ~to construct a \textbf{dataset of 12,820 (document, role, question-set) triples}.
    \item Second, we apply this dataset via fine-tuning and \textbf{demonstrate significant gains in zero-shot extraction}.
\end{enumerate}


\end{itemize}

\begin{figure*}[!htb]
  \centering
  \includegraphics[width =\linewidth]{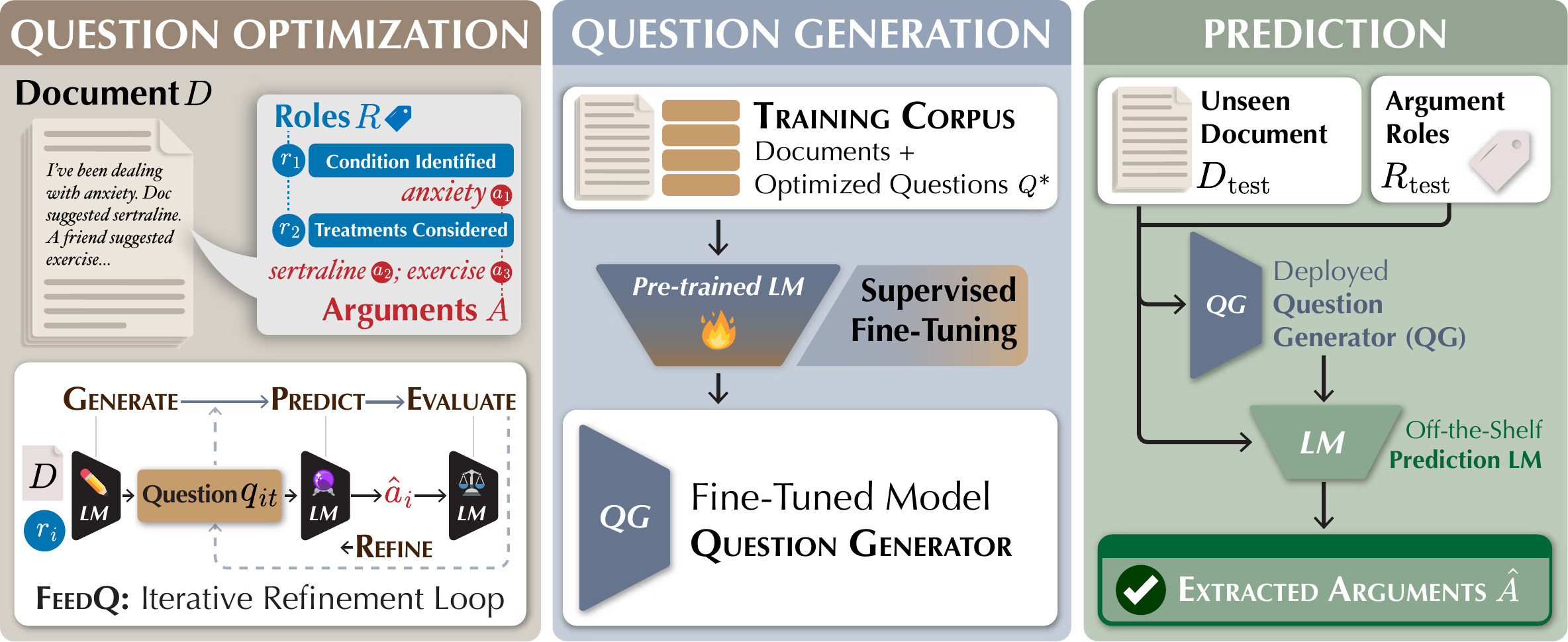}
 \caption{Overview of the List of Questions framework.
\textbf{(1)~Question optimization:} \FeedQ~iteratively
generates, evaluates, and refines role-specific questions
conditioned on the document, roles, and ground-truth arguments,  producing the right questions.
\textbf{(2)~QG fine-tuning:} These questions teach a smaller model to produce questions from documents
and roles, removing the dependence on ground-truth.
\textbf{(3)~Prediction:} At test time, the
fine-tuned QG model generates questions for unseen
document-roles, and a prediction model uses them to extract arguments. 
}
 \label{fig:loq-framework}
\end{figure*}

\section{Related Work}
\paragraph{Information extraction as question answering.}
Information extraction can be straightforwardly recast as question answering~\citep{du-cardie-2020-event,liu-etal-2020-event,li-etal-2020-event} by asking models a question whose answer is the desired argument. This prompts extraction models to use the question-answering and reading-comprehension abilities already present in pretrained LMs, and supports free-text rather than span-only approaches~\cite{huang-etal-2024-textee,sharif-etal-2024-explicit}. Once framed this way, question quality becomes instrumental to extraction success. In practice, questions are usually written once from annotation guidelines or role templates and then reused across all documents~\cite{hsu-etal-2022-degree,srivastava-etal-2025-instruction}. That makes them easy to apply, but also indifferent to how evidence is expressed in different documents.

\paragraph{Quality of questions.}
Recent work has started to take question formulation more seriously. \citet{lu-etal-2023-event}~generate better role questions from event-specific templates, \citet{hong-liu-2024-towards}~refine question generation with reinforcement learning, and \citet{uddin-etal-2024-generating}~show benefits from combining contextualized and uncontextualized questions. Despite this progress, most approaches still operate on a question-per-role paradigm~\citep{uddin-etal-2024-asking}, assuming each role is a single, clean retrieval target. However, a role may involve several arguments, be expressed indirectly, or scattered across sections. Current methods also usually improve question quality using proxies like template supervision~\citep{lu-etal-2023-event}, fluency and context-recovery rewards~\citep{hong-liu-2024-towards}, or heuristic combinations of question types~\citep{uddin-etal-2024-generating}. Our work builds on these ideas, but instead studies question design as a separable component of extraction. \textbf{We ask: for a document-role pair, what set of questions best exposes the relevant evidence and how can we find these?}

\paragraph{Learning to ask the right questions.}
\textit{Prompt optimization} is an emerging strategy for improving textual artifacts~\citep{pryzant2023automatic,yuksekgonul2025optimizing,lee2025feedback,cherep2026visualpersuasion}. \updt{Such techniques approximate gradient-based optimization via natural language \textit{feedback}. However, these methods optimize a single text prompt, whereas our setting requires optimizing a document- and role-conditioned \textit{set} of questions \textit{under a multi-tier extraction objective}, with set-level edits and a leakage constraint. We develop a new feedback-based optimizer which alternates question generation and information extraction, along with a multi-step evaluation and, document- and role-conditioning, a leakage check module, among several other components. We use this method to iteratively refine questions to maximize extraction quality.} We then transfer this knowledge via supervised fine-tuning to a lightweight question generation LM, which produces document-specific questions \textit{without} access to ground-truth arguments. Answering them with an off-the-shelf prediction LM should then yield, on average, better information extraction.

\section{Methods}


We propose the \FrameworkFull~(\textbf{\Framework}) framework, which automates the full argument extraction cycle in three phases (illustrated in Figure \ref{fig:loq-framework}).

\noindent
\textbf{Phase 1: Question Optimization.}
Given a document, role, and ground-truth arguments, we run an iterative optimization loop (\FeedQ, Section~\ref{sec:feedQ}) to derive questions that best extract those arguments, producing gold question sets for training.

\noindent
\textbf{Phase 2: Question Generation.}
We fine-tune a question generator (\QG) model on these gold questions, thereby teaching it to generate effective questions conditioned on document and role.

\noindent
\textbf{Phase 3: Prediction.}
At inference, the fine-tuned \QG~generates questions for unseen instances, and off-the-shelf LMs use these to extract arguments. Our design separates question generation from prediction, allowing each component to specialize. 

\subsection{Preliminaries}
\label{sec:preliminaries}
A \textbf{document} $\mathcal{D}$ is a full-length text taken from diverse clinical sources, such as case reports, pharmacovigilance records, or online health forums (e.g., reddit posts). Each $\mathcal{D}$ has an information structure defined by a set of \textbf{roles} $\mathcal{R} = \{r_1, r_2, \dots, r_n\}$ that specify the \textit{types} of information to extract. \textbf{Arguments} $\mathcal{A} = \{a_1, a_2, \dots, a_m\}$ are the corresponding details for each of these roles. For example, given the sentence \textit{``Patient was prescribed Metformin 500mg twice daily,''} the relevant information structure is a \textsc{Medication} event with roles \textit{drug}, \textit{dosage}, and \textit{frequency}, filled by the arguments \textit{Metformin}, \textit{500mg}, and \textit{twice daily}, respectively. We formulate clinical information extraction as a generative argument extraction task: given a document $\mathcal{D}$ and a set of target roles $\mathcal{R}$, an LM extracts a set of arguments $\mathcal{A}$ corresponding to each role.


\subsection{Question Optimization via Feedback}
\label{sec:feedQ}
We design \textbf{\FeedQ}, an iterative Feedback-Driven Question Optimization approach with the goal of automatically derive the right set of questions to extract the information from a given document.



\begin{algorithm}[!htb]
\small
\caption{\FeedQ: Feedback-Driven Question Optimization}\label{alg:feedq}
\begin{algorithmic}[1]
    \Require Ground-truth arguments $\mathcal{A}$, initial questions $\mathcal{Q}^{(0)}$, document $\mathcal{D}$, role $r$, max iterations $T$, target score $\tau$, patience $m$
    \Ensure Optimized question set $\mathcal{Q}^*$

    \State $\mathcal{Q}^* \leftarrow \mathcal{Q}^{(0)}$; \; $\tau^* \leftarrow 0.0$; \; $p \leftarrow 0$ \Comment{Initialization}

    \For{$t = 0, 1, \ldots, T{-}1$}
        \State $\hat{\mathcal{Q}}^{(t)} \leftarrow \textsc{LeakageCheck}(\mathcal{Q}^{(t)},\, \mathcal{A})$
        \State $\hat{\mathcal{A}}^{(t)} \leftarrow \textsc{Extract}(\mathcal{D},\, r,\, \hat{\mathcal{Q}}^{(t)})$ \Comment{Prediction}

        \Statex \quad \textit{// Evaluation}

        \State $M^{(t)} \leftarrow \hat{\mathcal{A}}^{(t)} \cap \mathcal{A}$
        \State $\text{Miss}^{(t)} \leftarrow \mathcal{A} \setminus \hat{\mathcal{A}}^{(t)}$
        \State $\text{Extra}^{(t)} \leftarrow \hat{\mathcal{A}}^{(t)} \setminus \mathcal{A}$

        \State $s^{(t)} \leftarrow E\left(|M^{(t)}|,\, |\hat{\mathcal{A}}^{(t)}|,\, |\mathcal{A}|\right)$ \Comment{Evaluation}

        \If{$s^{(t)} > \tau^*$} \Comment{Save best \& update patience}
            \State $\mathcal{Q}^* \leftarrow \hat{\mathcal{Q}}^{(t)}$
            \State $\tau^* \leftarrow s^{(t)}$
            \State $p \leftarrow 0$
        \Else
            \State $p \leftarrow p + 1$
        \EndIf

        \If{$\tau^* \geq \tau$ \textbf{or} $p \geq m$} \Comment{Early exit}
            \State \Break
        \EndIf

        \Statex \quad \textit{// Feedback signal creation}
        \State $\Delta^{(t)} \leftarrow \left(M^{(t)},\; \text{Miss}^{(t)},\; \text{Extra}^{(t)},\; s^{(t)},\; \hat{\mathcal{Q}}^{(t)}\right)$
        \Statex \quad \textit{// Question Refinement}
        \State $\mathcal{Q}^{(t+1)} \leftarrow \textsc{Refine}\left(\hat{\mathcal{Q}}^{(t)},\, \Delta^{(t)}\right)$
    \EndFor

    \State \Return $\mathcal{Q}^*$
\end{algorithmic}
\end{algorithm}

\paragraph{Objective.} Given a document $\mathcal{D}$, a role $r \in \mathcal{R}$, and
ground-truth arguments $\mathcal{A} = \{a_1, \ldots, a_m\}$ for $r$, we
seek a question set $\mathcal{Q}^*$ that maximizes extraction performance:
\begin{equation}
    \mathcal{Q}^* = \arg\max_{\mathcal{Q}} \;
    \text{F1}\big(\hat{\mathcal{A}}(\mathcal{D}, r, \mathcal{Q}),\;
    \mathcal{A}\big)
\end{equation}
where $\hat{\mathcal{A}}(\mathcal{D}, r, \mathcal{Q})$ denotes the
arguments extracted by a prediction model (\PredModel) conditioned on
$\mathcal{D}$, $r$, and $\mathcal{Q}$. We optimize $\mathcal{Q}$
via an iterative feedback loop (Algorithm~\ref{alg:feedq}) whose main components are:

\paragraph{Initialization.}
At $t{=}0$, the question generator (\QG) model produces an initial list of questions in a zero-shot manner conditioned on $r$, $\mathcal{D}$, and $\mathcal{A}$. 

\paragraph{Leakage Check.}
Since \QG~models can see $\mathcal{A}$, they could potentially leak ground-truth arguments $(a_i \in \mathcal{A})$ into the generated questions, directly or indirectly. To guard against this, we use an LM-based leakage check. At each iteration, $\operatorname{LeakageCheck}(\mathcal{Q}^{(t)}, \mathcal{A})$ flags questions that potentially \textit{leak} ground truth and rewrites them to remove the leak (e.g.\ asking a more generic question). We implement this module using \textsc{gpt-oss-120b}. Appendix~\ref{apdx:leakage-analysis} includes robustness analysis.

\paragraph{Extraction and Evaluation.}
Leakage-checked questions $\hat{\mathcal{Q}}^{(t)}$ are forwarded to the \PredModel, which outputs predicted arguments $\hat{\mathcal{A}}^{(t)}$. Predictions are evaluated against ground truth $\mathcal{A}$ using the evaluation framework (Section~\ref{sec:datasets-evaluation}). This yields sets of matched, missing, and over-generated arguments, from which we compute precision, recall, and F1.

\paragraph{Feedback and Refinement.} A structured feedback signal $\Delta$ at iteration $t$ contains the scores, matched, missed, and over-generated arguments, and passed to a refiner LM. Based on the current questions $\mathcal{Q}^{(t)}$ and feedback $\Delta^{(t)}$, the refiner generates an updated set $\mathcal{Q}^{(t+1)}$. The refinement is instructed to add questions targeting missed arguments, refine or remove questions causing over-generation, and preserve effective questions.

\paragraph{Convergence.}
We track the best question set $\mathcal{Q}^*$ across iterations by F1 score. The loop terminates when (i) F1 $\geq$ target threshold, (ii) no improvement for $p$ consecutive iterations (patience), or (iii) the maximum iteration count $T$ is reached. The output is always $\mathcal{Q}^*$ (best, not necessarily final since the F1 score is not guaranteed to be monotonic).

\subsection{Learning to Generate Questions}
\label{sec:question-generation}
\FeedQ~produces gold questions and, in doing so, yields a dataset of $(\mathcal{D}, r, \mathcal{Q}^*)$ triples. We use these triples to supervise the fine-tuning of Qwen3-4B and Qwen3-8B with LoRA. The resulting model learns to generate document- and role-conditioned questions without access to ground-truth, making \Framework~applicable to new queries at test time.

\subsection{Prediction}
\label{section:prediction-models}

At inference time, the fine-tuned \QG~model generates questions conditioned on the document and target roles. A prediction model then uses these questions to extract arguments from the document.  We evaluate five prediction models spanning both open-weight and proprietary LLMs: Qwen3-4B, Qwen3-8B, GPT-OSS-120B, GPT-5-Mini, and Gemini-3.1-Pro. Each \PredModel~is paired with every \QG, yielding a complete comparison across both axes. Details in Appendix \ref{apdx:models}, full prompts in Appendix \ref{section:prompts}.

\begin{table*}[!htb]
\centering
\small
\renewcommand*{\arraystretch}{1}
\begin{tabular}{l|C{0.7cm}C{0.7cm}C{0.7cm}C{0.7cm}C{0.8cm}c|C{0.7cm}C{0.7cm}C{0.7cm}C{0.7cm}C{0.8cm}c}
 \rowcolor{cyan!10}& Qw3 4B & Qw3 8B & OSS 120B & Gpt 5 mini & Gemini 3.1 & \textbf{Mean} & Qw3 4B & Qw3 8B & OSS 120B & Gpt 5 mini & Gemini 3.1 & \textbf{Mean} \\
\midrule
\textbf{Approach} & \multicolumn{6}{c}{\textit{\textbf{CaseReportBench}}} & \multicolumn{6}{c}{\textit{\textbf{PHEE}}} \\
\midrule
\updt{No-Question}  & 55.6 & 54.9 & 70.1 & 65.2 & 70.6 & 63.3 & 62.8 &	67.8 & 66.2 & 66.4	& 83.3  & 69.3\\

\midrule
\multicolumn{13}{c}{\textit{Performance with Knowledge Questions}}\\
\midrule
 Knowledge-Q & 62.4 & 61.6 & 75.3 & 71.0 & 78.3 & 69.7 & 76.5 & 79.5 & 75.7 & 71.8 & 83.4 & 77.4 \\
 CoT-Q & 62.8 & 57.5 & 73.9 & 67.9 & 77.2 & 67.9 & 69.2 & 73.5 & 76.2 & 78.6 & 84.2 & 76.3 \\
 \midrule
\multicolumn{13}{c}{\textit{Performance with Dynamic Questions generated with GPT-OSS-120B}}\\
\midrule
 Contextual-Q & 54.3 & 50.6 & 57.9 & 52.5 & 57.7 & 54.6 & 49.3 & 47.5 & 51.0 & 52.9 & 79.5 & 56.0 \\
 Zs-\FeedQ & 76.9 & 72.1 & 92.7 & 88.7 & 91.9 & 84.5 & 82.5 & 83.5 & 86.3 & 84.4 & 89.4 & 85.2 \\
\rowcolor{green!25}  Opt-\FeedQ & 79.2 & 73.4 & 96.9 & 92.4 & 94.0 & 87.2 & 82.3 & 84.8 & 94.3 & 89.8 & 90.6 & 88.4 \\
\midrule
\midrule
& \multicolumn{6}{c}{\textit{\textbf{DiscourseEE}}} & \multicolumn{6}{c}{\textit{\textbf{MACCROBAT}}} \\
\midrule
\updt{No-Question}  & 46.6 &	46.1 & 	47.2 &	44.9 &	52.2 &	47.4 &	30.3 &	33.2 &	37.1 &	37.1 &	52.0 &	37.9 \\
\midrule
\multicolumn{13}{c}{\textit{Performance with Knowledge Questions}}\\
\midrule
 Knowledge-Q & 53.0 & 53.8 & 60.5 & 55.6 & 57.8 & 56.1 & 36.6 & 38.3 & 42.1 & 42.5 & 58.1 & 43.5 \\
 CoT-Q & 54.7 & 55.7 & 61.9 & 53.6 & 60.1 & 57.2 & 36.9 & 38.6 & 42.5 & 40.0 & 57.1 & 43.0 \\
\midrule
\multicolumn{13}{c}{\textit{Performance with Dynamic Questions generated with GPT-OSS-120B}}\\
\midrule
 Contextual-Q & 42.2 & 41.3 & 43.5 & 41.0 & 40.3 & 41.7 & 30.5 & 31.3 & 30.5 & 29.3 & 38.3 & 32.0 \\
 Zs-\FeedQ & 60.0 & 61.2 & 72.2 & 61.2 & 67.3 & 64.4 & 49.4 & 56.3 & 74.9 & 64.1 & 83.6 & 65.6 \\
 \rowcolor{green!25} Opt-\FeedQ & 65.5 & 64.8 & 82.5 & 64.6 & 69.0 & 69.3 & 57.3 & 66.5 & 91.1 & 74.6 & 91.3 & 76.2 \\
\bottomrule
\end{tabular}
\caption{Impact of question quality on dev set extraction performance across four datasets and five prediction models. For each column, only the questions change, isolating the effect of asking the right questions. Zs-\FeedQ\ are questions generated in one zero-shot pass and Opt-\FeedQ\ are questions iteratively refined with \FeedQ.}
\label{table:dev_results_cm}
\end{table*}

\section{Datasets and Evaluation}
\label{sec:datasets-evaluation}
We evaluate on four clinical and biomedical IE datasets spanning different text types and annotation granularities:
\textbf{CaseReportBench}~\cite{pmlr-v287-zhang25b}, comprising full-length clinical case reports (avg.\ 582 words); \textbf{PHEE}~\cite{sun-etal-2022-phee}, consisting of short pharmacovigilance texts (avg.\ 20 words); \textbf{DiscourseEE}~\cite{sharif-etal-2024-explicit}, drawn from informal online health forums; and \textbf{MACCROBAT}~\cite{ma-etal-2023-dice}, containing PubMed case report snippets.
All datasets are transformed into the unified format described in Section~\ref{sec:preliminaries}: triplets of (document, role, list of ground-truth arguments). Data statistics are given in Table~\ref{table:dataset-statistics}.

Each dataset is divided into training, development, and test splits. The development set is used to design and internally validate \FeedQ. The training set provides the samples on which \FeedQ~produces gold questions for fine-tuning. \textbf{The test set is held out for final evaluation on unseen data}. More details are in Appendix~\ref{apx:dataset-details}. We report F1 for all approaches on test set. Exact-match and relaxed-match metrics are known to underestimate LLM performance, as semantically correct predictions with surface-form differences are penalized \cite{lu2025exactmatchsemanticallyreassessing,fane-etal-2025-bemeae}.
We adopt the hierarchical approach of \citet{sharif-etal-2025-regen}, which combines exact, relaxed, and LLM-as-judge to evaluate argument correctness (details in Appendix \ref{apdx:evaluation}).




\section{Importance of Better Questions}
\label{section:asking-right-Q}

To first \textbf{isolate the effect of question quality} on extraction, we compare approaches that vary only how questions are formed while holding the prediction model fixed (Table~\ref{table:dev_results_cm}). \updt{In No-Question, we prompt the prediction model with (document, role) and instruct it to generate arguments without questions. \textbf{Performance degrades across all datasets, establishing that questions are necessary}. Knowledge-Q and CoT-Q are \textit{knowledge questions}: defined once per role and applied uniformly across all documents \cite{du-cardie-2020-event,hsu-etal-2022-degree}. These are directly taken from expert questions released with the dataset or minimally adapted to QA format from their expert annotation guidelines (details: Appendix \ref{knowledge-q-source}). These are a strong human-written question baseline.} Contextual-Q generates \textit{dynamic questions} per document–role pair \textit{without} access to ground truth, a realistic deployment setting \cite{hong-liu-2024-towards}. \updt{Zs-\FeedQName~and Optimized-\FeedQName~also generate dynamic questions, but initially \textit{with} access to ground-truth arguments: the former in a single zero-shot pass, the latter through iterative refinement via \FeedQ.} Ground-truth access \textbf{establishes an upper bound on what is achievable when questions are well-targeted}, before asking whether that quality can be reproduced without it. For experiments, we use the prediction models discussed in Section \ref{section:prediction-models}.

One key observation in Table \ref{table:dev_results_cm} is that Contextual-Q degrades below No-Question baseline. This raises the question of why document-condition questions hurt the extraction performance. To investigate this, we did a precision-recall analysis, shown in Table \ref{table:pr}. We find that \textbf{unoptimized Contextual Questions lead the prediction models to generate many extra candidate arguments, inflating false positives.}

\begin{table}[!htb]
\small
\renewcommand*{\arraystretch}{0.95}
\centering
\begin{tabular}{ll|ccc}
\toprule
\textbf{Dataset} & \textbf{Approach} & \textbf{P} & \textbf{R} & \textbf{F1} \\
\midrule

\multirow{2}{*}{CaseReportB}
& No-Question  & 54.5 & 77.5 & 63.3 \\
& Contextual-Q & 40.2 & 85.8 & 54.6 \\
\midrule

\multirow{2}{*}{PHEE}
& No-Question  & 63.6 & 76.9 & 69.4 \\
& Contextual-Q & 44.6 & 78.6 & 56.0 \\
\midrule

\multirow{2}{*}{DiscourseEE}
& No-Question  & 37.4 & 65.7 & 47.4 \\
& Contextual-Q & 28.8 & 76.3 & 41.7 \\
\midrule

\multirow{2}{*}{MACCROBAT}
& No-Question  & 29.1 & 54.8 & 38.0 \\
& Contextual-Q & 22.1 & 57.9 & 32.0 \\
\bottomrule

\end{tabular}
\caption{Overall precision, recall, and F1 analysis of the No-Question and Contextual-Q approaches.}
\label{table:pr}
\end{table}

Both No-Question and Contextual-Q condition on the document and the target role; Contextual-Q additionally provides questions generated on the fly without ground-truth access. \textbf{It generates 5.27 questions per document–role pair on average which causes the trade-off above}. Across every dataset it raises recall (e.g., DiscourseEE 65.7 → 76.3) but lowers precision by large margin (37.4 → 28.8). In contrast, No-question produces fewer arguments, so its recall is lower. However, large prediction models can still recover many common arguments directly from the document and role specification, resulting in substantially higher precision and, ultimately, higher F1.


\subsection{Discovering the Right Questions} 
We select the best questions using \FeedQ. Selecting a question generator for \FeedQ~requires balancing quality and cost. We evaluate five LLMs as zero-shot question generators (Qwen3-4B, Qwen3-8B, GPT-OSS-120B, GPT-5.2, and Gemini-3.1-Pro), the latter two with high reasoning. For each, we compute the F1 score averaged across all prediction models. Table \ref{table:qg_ablation} summarizes the results; per-model breakdown is illustrated in Figure ~\ref{figure:qg-ablation}.

\begin{table}[!htb]
\small
\renewcommand*{\arraystretch}{1}
\centering
\begin{tabular}{l|cccc|c}

 \rowcolor{cyan!10}\textbf{QG} & CRB & PHEE & DEE & MBAT & \textbf{Mean} \\
\midrule
Qwen3-4B          & 60.8 & 66.8 & 48.3 & 43.3 & \textbf{54.8} \\
Qwen3-8B          & 60.8 & 64.1 & 47.8 & 40.7 & \textbf{53.4} \\
OSS-120B      & 84.5 & 85.2 & 64.4 & 65.6 & \textbf{74.9} \\
GPT-5.2      & 85.1 & 85.9 & 65.8 & 67.8 & \textbf{76.1} \\
Gemini-3.1 & 79.8 & 85.1 & 63.5 & 59.8 & \textbf{72.1} \\
\bottomrule
\end{tabular}
\caption{Zero-shot question generation quality across five LLMs, measured by F1 averaged over prediction models. Per-model scores shown in Figure~\ref{figure:qg-ablation}. CRB, DEE, and MBAT indicate CaseReportBench, DiscourseEE, and MACCROBAT datasets, respectively.}
\label{table:qg_ablation}
\end{table}


Larger models outperform the Qwen3 baselines by a wide margin ($\approx$72--76 vs. $\approx$54 F1). GPT-5.2 performs best overall (76.1), but GPT-OSS-120B is close (74.9) while being over 40$\times$ cheaper per token at the time of writing. We therefore use GPT-OSS-120B as the \QG~model. Its open weights additionally allow local deployment and use with private data when needed. For scalability, we also use GPT-OSS-120B as the prediction and refinement model, though any sufficiently capable model could be substituted.

\begin{table*}[!htb]
\centering
\small
\renewcommand*{\arraystretch}{0.92}
\begin{tabular}{l|C{1cm}C{1cm}C{0.8cm}C{0.8cm}C{1cm}|C{1cm}C{1cm}C{0.8cm}C{0.8cm}C{1cm}}
  \rowcolor{cyan!10}& Qwen3 4B & Qwen3 8B & OSS 120B & Gpt-5 mini & Gemini 3.1-pro & Qwen3 4B & Qwen3 8B & OSS 120B & Gpt-5 mini & Gemini 3.1-pro \\
\midrule
\textbf{Approach} & \multicolumn{5}{c|}{\textit{\textbf{CaseReportBench}}} & \multicolumn{5}{c}{\textit{\textbf{PHEE}}} \\

\midrule
\multicolumn{11}{c}{\textit{Performance with Knowledge Questions}}\\
\midrule
 Knowledge-Q & 60.29 & 60.17 & \cellcolor{green!15}73.40 & \cellcolor{green!15}69.03 & \cellcolor{green!50}78.17 & 76.92 & 80.02 & 74.08 & 74.05 & 81.37 \\
 CoT-Q & 57.98 & 58.24 & \cellcolor{green!50}73.92 & 66.33 & \cellcolor{green!15}76.32 & 67.98 & 73.67 & 77.50 & 76.41 & 80.53 \\
\midrule
\multicolumn{11}{c}{\textit{Performance with fine-tuned and non fine-tuned QG models}}\\
\midrule
Qwen3-4B & 63.18 & 59.26 & 52.28 & 45.73 & 66.14 & 64.71 & 63.54 & 62.30 & 58.75 & 79.89 \\
Qwen3-8B & \cellcolor{green!15}63.26 & 60.51 & 56.67 & 49.71 & 67.33 & 62.10 & 63.26 & 58.58 & 55.42 & 79.11 \\
\multicolumn{11}{c}{- - - - - - - - - - - - - - - - - - - - - - - - - - - - - - - - - - - - - - - - - - - - - - - - - - - - - - - - - - - - - - - - - - - - - - - - - - - - - - - - - - } \\

Qwen3-4B$^{\text{ft*}}$ & 62.56 & \cellcolor{green!15}\textbf{64.09} & 71.63 & 67.86 & 71.32 & \cellcolor{green!15}78.48 & \cellcolor{green!50}\textbf{81.52} & \cellcolor{green!15}\textbf{83.61} & \cellcolor{green!15}\textbf{80.30} & \cellcolor{green!50}\textbf{84.96} \\


Qwen3-8B$^{\text{ft*}}$ & \cellcolor{green!50}\textbf{63.85} & \cellcolor{green!50}\textbf{64.59} & 70.88 & \cellcolor{green!50}\textbf{69.16} & 69.98 & 78.07 & 80.66 & \cellcolor{green!50}\textbf{84.52} & \cellcolor{green!50}\textbf{82.74} & \cellcolor{green!15}\textbf{84.25} \\
\midrule
\multicolumn{11}{c}{\textit{Strong QG Baselines (non fine-tuned)}}\\
\midrule
Gpt-OSS-120B & 53.76 & 49.45 & 56.31 & 51.14 & 55.88 & 48.57 & 44.87 & 49.40 & 49.81 & 77.35 \\
Gpt-5.2 & 60.46 & 58.07 & 69.36 & 64.13 & 65.63 & 62.59 & 61.39 & 71.87 & 71.44 & 80.33 \\
Gemini-3.1 & 62.01 & 59.39 & 70.21 & 67.04 & 71.26 & \cellcolor{green!50}79.39 & \cellcolor{green!15}80.68 & 80.98 & 80.13 & 81.89 \\
\midrule
\midrule
 & \multicolumn{5}{c|}{\textit{\textbf{DiscourseEE}}} & \multicolumn{5}{c}{\textit{\textbf{MACCROBAT}}} \\
\midrule
\multicolumn{11}{c}{\textit{Performance with Knowledge Questions}}\\
\midrule
Knowledge-Q & \cellcolor{green!15}50.55 & \cellcolor{green!15}53.68 & \cellcolor{green!15}60.36 & \cellcolor{green!50}55.15 & \cellcolor{green!15}58.13 & 37.55 & 39.11 & 43.70 & 41.36 & \cellcolor{green!50}58.74 \\
 CoT-Q & 50.44 & \cellcolor{green!50}55.13 & \cellcolor{green!50}61.01 & 52.71 & \cellcolor{green!50}59.62 & 37.41 & 41.75 & 43.60 & 41.60 & \cellcolor{green!15}58.51 \\
\midrule
\multicolumn{11}{c}{\textit{Performance with fine-tuned and non fine-tuned QG models}}\\
\midrule
Qwen3-4B & 50.11 & 48.75 & 47.13 & 44.96 & 47.64 & 38.02 & 41.50 & 41.07 & 38.72 & 48.43 \\
Qwen3-8B & 47.91 & 48.90 & 46.59 & 42.66 & 44.36 & 38.71 & 40.28 & 40.12 & 39.07 & 47.57 \\
\multicolumn{11}{c}{- - - - - - - - - - - - - - - - - - - - - - - - - - - - - - - - - - - - - - - - - - - - - - - - - - - - - - - - - - - - - - - - - - - - - - - - - - - - - - - - - - } \\

Qwen3-4B$^{\text{ft*}}$ & 49.75 & 50.16 & 54.12 & 54.01 & 51.46 & \cellcolor{green!50}\textbf{45.37} & \cellcolor{green!50}\textbf{46.37} & \cellcolor{green!50}\textbf{51.39} & \cellcolor{green!50}\textbf{48.62} & 56.84 \\
Qwen3-8B$^{\text{ft*}}$ & 47.93 & 50.57 & 56.74 & \cellcolor{green!15}54.47 & 52.71 & \cellcolor{green!15}\textbf{41.81} & \cellcolor{green!15}\textbf{44.96} & \cellcolor{green!15}\textbf{50.96} & \cellcolor{green!15}\textbf{47.53} & 56.74 \\
\midrule
\multicolumn{11}{c}{\textit{Strong QG Baselines (non fine-tuned)}}\\
\midrule
Gpt-OSS-120B & 41.66 & 40.52 & 44.60 & 41.59 & 38.83 & 32.49 & 32.51 & 33.31 & 32.90 & 39.18 \\
Gpt-5.2 & 48.89 & 43.95 & 50.55 & 46.65 & 41.04 & 39.57 & 39.88 & 40.50 & 39.30 & 42.91 \\
Gemini-3.1 & \cellcolor{green!50}51.39 & 49.65 & 53.18 & 50.84 & 49.52 & 37.13 & 40.59 & 43.25 & 41.49 & 52.13 \\
\bottomrule
\end{tabular}
\caption{Test-set performance comparison across question generation approaches. Knowledge-Q and CoT-Q are human-written questions. All other approaches generate questions dynamically, conditioned only on the document and role (no ground-truth access). Qwen3\textsuperscript{ft} rows report the best score after fine-tuning (see Table \ref{table:ft_ablation} for the full breakdown). Bold indicates the fine-tuned model beats all approaches for that dataset; \colorbox{green!50}{green}  and \colorbox{green!15}{light green} mark the first and second highest scores overall.}
\label{table:test_results_cm}
\end{table*}

\subsection{Impact of Question Improvement}
Better questions would yield only marginal gains if the bottleneck in LLM-based extraction were model capability. Table~\ref{table:dev_results_cm} results suggest otherwise. Optimized-\FeedQName~outperforms the strongest baseline (Knowledge-Q) by \textbf{18.6 F1} averaged across four datasets and all prediction models. Gains hold on every dataset–model combination, ranging from 11.0 (PHEE) to 32.7 (MACCROBAT).

Zero-shot \FeedQName~already exceeds Knowledge-Q by 13.2 F1 without iterative refinement, indicating that much of the gain comes from generating questions informed by what information the document actually contains for a given role. \FeedQName~optimization adds a further 5.4 F1, with the largest gain on DiscourseEE (+4.9) and MACCROBAT (+10.6), where arguments are most frequently implicit or distributed. Neither alternative questioning strategy helps: Contextual-Q falls 15.6 F1 below Knowledge-Q despite conditioning on the document, and CoT-Q shows no benefit either (61.1 vs.\ 61.7). \textbf{Document conditioning without grounding produces vague queries and reasoning does not compensate for a poorly targeted question.}

\begin{figure}[!htb]
  \centering
    \includegraphics[width=\linewidth]{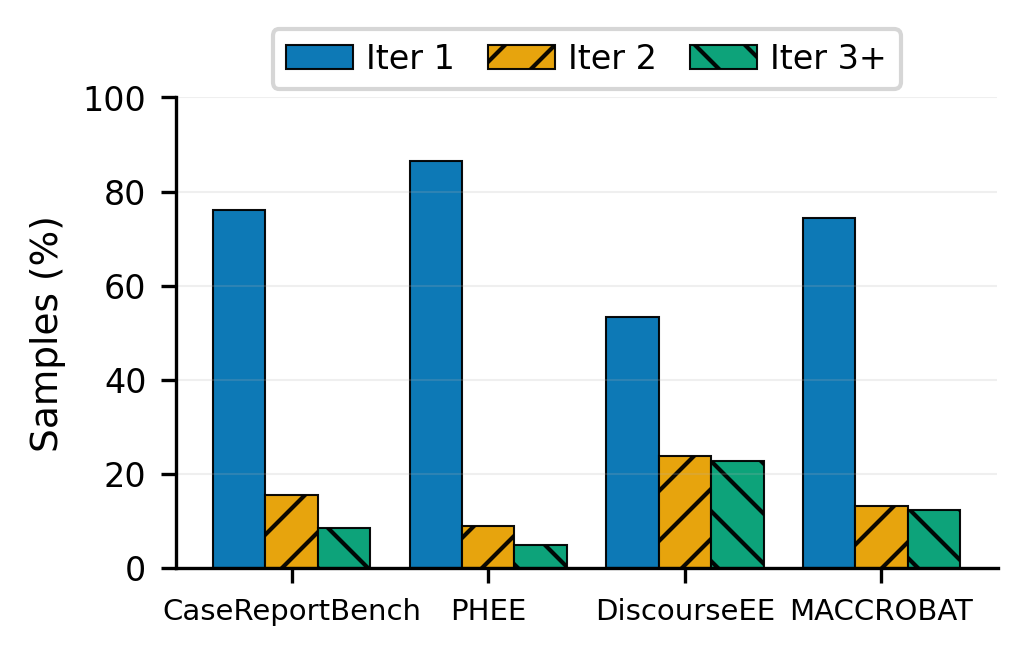}
    \caption{Distribution of the \FeedQ~iteration at which the best-scoring question set is found in the dev set.}
    \label{fig:iter_dist}
\end{figure}

\begin{figure}[!htb]
    \includegraphics[width=\linewidth]{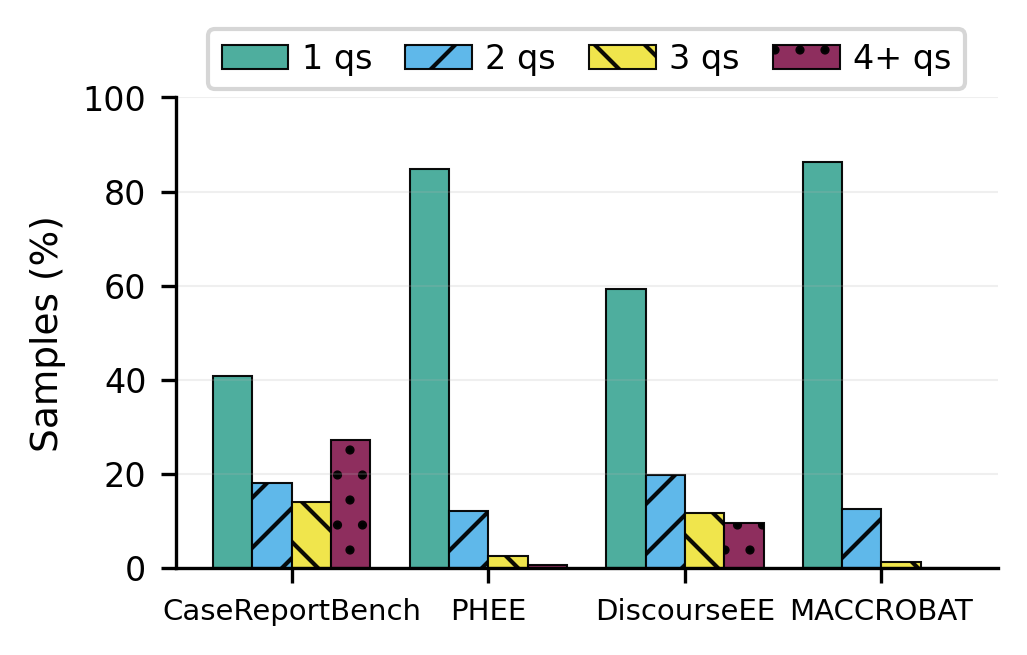}
    \caption{Distribution of the percentage of optimized questions generated per role in the development set.}
    \label{fig:qcount_dist}
\end{figure}

\subsection{Analysis of Optimized Questions}
Figure \ref{fig:iter_dist} and \ref{fig:qcount_dist} provide justification of two core design choices in our pipeline. First, iterative refinement is not redundant: while a single \FeedQ~pass suffices for the majority of roles in PHEE (87\%) and MACCROBAT (75\%), nearly half of DiscourseEE roles (47\%) and a quarter of CaseReportBench roles require two or more iterations before reaching their best-scoring question set (Figure~\ref{fig:iter_dist}). Second, a single question per role is frequently insufficient. CaseReportBench assigns only 41\% of roles a single optimized question; the remaining 59\% receive two or more, with 27\% requiring four or more (Figure~\ref{fig:qcount_dist}). DiscourseEE follows a similar pattern. This confirms that roles in structurally rich schemas often demand multiple targeted questions to adequately decompose the extraction target. For PHEE and MACCROBAT, single well-phrased questions work well enough in the majority of cases. In Appendix~\ref{section:interp}, we add supplementary evidence showing \textit{how} the questions change using auto-interpretability methods.

\section{Learning the Right Questions}

Section~\ref{section:asking-right-Q} established that \textbf{better questions improve extraction}. The practical question is whether a model can learn to generate effective questions for \textit{unseen} queries, i.e. without access to ground-truth arguments. We test this by fine-tuning Qwen3-4B and Qwen3-8B models on the questions produced by \FeedQ. Specifically, we run \FeedQ~on the training split of each dataset to generate (document, role, optimized question-set) triples, producing \textbf{12,820 triples} in total. We then perform supervised fine-tuning with LoRA to train each model to map a document and role to a question (Section \ref{sec:question-generation}). We construct three training configurations: a \textit{balanced mix} that caps each dataset at a comparable sample count ($\approx$6K total), an \textit{all-available} mix that uses every training sample ($\approx$12K total), and an \textit{in-domain} mix that trains on only the target dataset's triples. We report the best of the three per model and provide all comparisons in Table~\ref{table:ft_ablation}. More fine-tuning details are in Appendix \ref{sec:fine-tuning-details}.

\begin{table}[!htb]
\small
\renewcommand*{\arraystretch}{1}
\centering
\begin{tabular}{l|cccc|c}
\rowcolor{cyan!10}\textbf{Approach} & CRB & PHEE & DEE & MBAT & Mean \\
\midrule
\multicolumn{6}{c}{\textit{Knowledge Questions}}\\
\midrule
Knowledge-Q         & \cellcolor{green!50}68.2 & 77.3 & \cellcolor{green!15}55.6 & 44.1 & 61.3 \\
CoT-Q            & 66.6 & 75.2 & \cellcolor{green!50}55.8 & 44.6 & 60.5 \\
\midrule
\multicolumn{6}{c}{\textit{Fine-tuned and Non Fine-tuned QG Models}}\\
\midrule
Qwen3-4B           & 57.3 & 65.8 & 47.7 & 41.5 & 53.1 \\
Qwen3-8B           & 59.5 & 63.7 & 46.1 & 41.1 & 52.6 \\
\multicolumn{6}{c}{- - - - - - - - - - - - - - - - - - - - - - - - - - - - - - - - - - - - - - - } \\
Qwen3-4B$^{\text{ft-a}}$ & 67.5 & 80.7 & 51.9 & 49.0 & 62.3 \\
Qwen3-4B$^{\text{ft*}}$ & 67.5 & \cellcolor{green!15}\textbf{81.8} & 51.9 & \cellcolor{green!50}\textbf{49.7} & \cellcolor{green!50}\textbf{62.7} \\
Qwen3-8B$^{\text{ft-a}}$ & 64.7 & 81.6 & 51.5 & 47.2 & 61.3 \\
Qwen3-8B$^{\text{ft*}}$ & \cellcolor{green!15}67.7 & \cellcolor{green!50}\textbf{82.0} & 52.5 & \cellcolor{green!15}\textbf{48.4} & \cellcolor{green!15}\textbf{62.7} \\
\midrule
\multicolumn{6}{c}{\textit{Strong QG Baselines (Non Fine-tuned)}}\\
\midrule
OSS-120B       & 53.3 & 54.0 & 41.4 & 34.1 & 45.7 \\
Gpt-5.2            & 63.5 & 69.5 & 46.2 & 40.4 & 54.9 \\
Gemini-3.1         & 66.0 & 80.6 & 50.9 & 42.9 & 60.1 \\
\bottomrule
\end{tabular}
\caption{Test-set F1 averaged across five prediction models. Fine-tuned Qwen3 models (ft) are trained on optimized questions from \FeedQ; (ft* is best configuration per-model, ft-a is mixture-trained; full breakdown in Table~\ref{table:ft_ablation}). CRB, DEE, and MBAT indicate CaseReportBench, DiscourseEE, and MACCROBAT, respectively.
}
\label{table:test_avg_cm}
\end{table}

\updt{We compare against three groups of approaches. First, the Knowledge-Q and CoT-Q which require human written questions (Section \ref{section:asking-right-Q}).} Second, non-fine-tuned Qwen3-4B and Qwen3-8B generating questions for each document–role pair via contextual prompting (conditioned on document and role, no ground truth), isolating the effect of fine-tuning within the same models. Third, three strong non-fine-tuned \QG~models (GPT-OSS-120B, GPT-5.2, and Gemini-3.1-Pro) using the same contextual prompt, establishing what scale alone can achieve. \textbf{All results are reported on the held-out test set.} Table \ref{table:test_results_cm} gives per-model scores and Table \ref{table:test_avg_cm} averages across prediction models.

\subsection{Tuned models generate effective questions}

Fine-tuning on \FeedQ-generated data substantially improves question generation quality. After tuning, both Qwen3 models reach 62.7 mean F1, up from $\approx$53 without fine-tuning (Table~\ref{table:test_avg_cm}). More importantly, the fine-tuned models outperform all non-fine-tuned strong baselines, including GPT-OSS-120B, GPT-5.2, and Gemini-3.1. The largest gains appear on PHEE (+$\approx$16--18) and MACCROBAT (+$\approx$7-8), but the overall improvement is not dataset-specific. These results suggest that effective question generation depends largely on implementing the right questioning behavior.

Fine-tuned models beat Knowledge-Q on average (62.7 vs.\ 61.3). On PHEE and MACCROBAT, fine-tuned models outperform Knowledge-Q by 4.5--4.7 and 4.3--5.6 F1 respectively. On CaseReportBench, Knowledge-Q retains a marginal edge (under 1 F1), while on DiscourseEE the gap is larger (3--4 F1). This aligns with training data availability: PHEE and MACCROBAT each contribute 5,000 training samples, while CaseReportBench provides only 620 and DiscourseEE 2,200. With limited supervision, the fine-tuned models cannot fully learn the question patterns needed for these datasets. \updt{We provide systematic data scaling experiments in Appendix \ref{dataset-scaling-analysis}.}

\subsection{Out-of-domain Experiments}
\updt{We run additional zero-shot out-of-domain experiments on a subset of 500 samples on datasets: DocEE~\cite{tong-etal-2022-docee}, which has long, argument-dense tasks; GENEVA~\cite{parekh-etal-2023-geneva} which has brief, simple examples; and MUC4~\cite{muc-1992-message}, news documents. Overall, we find that fine-tuning benefits DocEE (the more long and complex samples), MUC4 but not GENEVA, which annotates multiple coreferent arguments per role and is better served by the broader question sets of the base model. Precision-recall comparison indicates that the fine-tuned model produces role-targeted questions that improve precision across all datasets. Detailed analysis is in Appendix~\ref{addtional-ood-analysis}.}


\begin{table}[!htb]
\centering
\small
\renewcommand*{\arraystretch}{1}
\setlength{\tabcolsep}{4pt}
\begin{tabular}{lcccccc}
\toprule
\rowcolor{cyan!10}
 & \multicolumn{3}{c}{\textbf{Qwen3-4B}} & \multicolumn{3}{c}{\textbf{Qwen3-8B}} \\
\cmidrule(lr){2-4} \cmidrule(lr){5-7}
\rowcolor{cyan!10}
\textbf{Dataset} & \textbf{P} & \textbf{R} & \textbf{F1} & \textbf{P} & \textbf{R} & \textbf{F1} \\
\midrule
\multicolumn{7}{l}{\textit{Base}} \\
DocEE  & 33.35 & 82.88 & 46.53 & 35.58 & 83.82 & 49.20 \\
GENEVA & 48.33 & 77.51 & \textbf{59.13} & 47.38 & 77.66 & 58.30 \\
MUC4   & 45.98 & 80.20 & 57.45 & 47.21 & 80.14 & 58.56 \\
\midrule
\multicolumn{7}{l}{\textit{Fine-tuned (ft*)}} \\
DocEE  & 49.18 & 63.63 & 55.28 & 48.38 & 68.65 & \textbf{56.65} \\
GENEVA & 49.73 & 56.43 & 52.82 & 49.66 & 56.65 & 52.88 \\
MUC4   & 53.82 & 61.86 & 57.30 & 54.34 & 67.24 & \textbf{59.90} \\
\bottomrule
\end{tabular}
\caption{\updt{Out-of-domain results on DocEE, GENEVA, and MUC4, averaged across all evaluator models (ft* is best configuration per model; full breakdown in Table~\ref{table:ood_results_full}). The highest F1 per dataset is bolded.}}
\label{table:ood_results_avg}
\end{table}

\section{Conclusion}
Extraction is a central ambition of information systems: how to find the right fact or passage when needed. However, documents are not databases; the information is not neatly organized awaiting extraction, and complex extractions involve representations of the problem and the data available to solve it. The query is inevitably part of this representation, and choosing it poorly can thus put an extractor on the wrong trail and render the necessary information invisible. In this paper, we have offered evidence of this problem and derived a scalable and automated solution. We invite the information extraction research community to draw on our methods and data to build better forms of information access for modern language technologies.

\section*{Limitations}
One limitation is that \FeedQ{} uses GPT-OSS-120B as the sole model for question generation, refinement, leakage checking, and prediction during optimization. The optimized questions transfer effectively across all five prediction models at test time (Tables \ref{table:test_results_cm}–\ref{table:test_avg_cm}), indicating that the learned patterns generalize beyond the optimizer. Exploring whether varying the optimizer or ensembling questions from multiple models yields further gains could yield valuable additional design knowledge.

\updt{Separately, all four primary benchmarks in this work are clinical and biomedical, where arguments are distributed across narrative sections and expressed in domain-specific terms. These properties that make targeted questioning particularly impactful. To further assess generalizability, we experimented with three out-of-domain non-clinical datasets. Results indicate that \Framework~remains effective in these settings. Extending \Framework~to other extraction domains such as news, law, or finance would demonstrate broader applicability.}

\section*{Reproducibility}
The datasets, models, and implementation details are provided in Appendices~\ref{apx:dataset-details}, \ref{apx:implementaion-details}, and \ref{section:prompts}. Our code and additional resources are available at \url{https://omar-sharif.github.io/LoQ}.
\bibliography{acl_latex}

@inproceedings{sharif-etal-2025-regen,
    title = "{REG}en: A Reliable Evaluation Framework for Generative Event Argument Extraction",
    author = "Sharif, Omar  and
      Gatto, Joseph  and
      Basak, Madhusudan  and
      Preum, Sarah Masud",
    editor = "Christodoulopoulos, Christos  and
      Chakraborty, Tanmoy  and
      Rose, Carolyn  and
      Peng, Violet",
    booktitle = "Findings of the Association for Computational Linguistics: EMNLP 2025",
    month = nov,
    year = "2025",
    address = "Suzhou, China",
    publisher = "Association for Computational Linguistics",
    url = "https://aclanthology.org/2025.findings-emnlp.649/",
    doi = "10.18653/v1/2025.findings-emnlp.649",
    pages = "12146--12168",
    ISBN = "979-8-89176-335-7",
}

@InProceedings{pmlr-v287-zhang25b,
  title = 	 {CaseReportBench: An LLM Benchmark Dataset for Dense Information Extraction in Clinical Case Reports},
  author =       {Zhang, Xiao Yu Cindy and Ferreira, Carlos R. and Rossignol, Francis and Ng, Raymond T. and Wasserman, Wyeth and Zhu, Jian},
  booktitle = 	 {Proceedings of the sixth Conference on Health, Inference, and Learning},
  pages = 	 {527--542},
  year = 	 {2025},
  editor = 	 {Xu, Xuhai Orson and Choi, Edward and Singhal, Pankhuri and Gerych, Walter and Tang, Shengpu and Agrawal, Monica and Subbaswamy, Adarsh and Sizikova, Elena and Dunn, Jessilyn and Daneshjou, Roxana and Sarker, Tasmie and McDermott, Matthew and Chen, Irene},
  volume = 	 {287},
  series = 	 {Proceedings of Machine Learning Research},
  month = 	 {25--27 Jun},
  publisher =    {PMLR},
  url = 	 {https://proceedings.mlr.press/v287/zhang25b.html},
}

@inproceedings{sharif-etal-2024-explicit,
    title = "Explicit, Implicit, and Scattered: Revisiting Event Extraction to Capture Complex Arguments",
    author = "Sharif, Omar  and
      Gatto, Joseph  and
      Basak, Madhusudan  and
      Preum, Sarah M.",
    editor = "Al-Onaizan, Yaser  and
      Bansal, Mohit  and
      Chen, Yun-Nung",
    booktitle = "Proceedings of the 2024 Conference on Empirical Methods in Natural Language Processing",
    month = nov,
    year = "2024",
    address = "Miami, Florida, USA",
    publisher = "Association for Computational Linguistics",
    url = "https://aclanthology.org/2024.emnlp-main.673/",
    doi = "10.18653/v1/2024.emnlp-main.673",
    pages = "12061--12081",
}

@inproceedings{sun-etal-2022-phee,
    title = "{PHEE}: A Dataset for Pharmacovigilance Event Extraction from Text",
    author = "Sun, Zhaoyue  and
      Li, Jiazheng  and
      Pergola, Gabriele  and
      Wallace, Byron  and
      John, Bino  and
      Greene, Nigel  and
      Kim, Joseph  and
      He, Yulan",
    editor = "Goldberg, Yoav  and
      Kozareva, Zornitsa  and
      Zhang, Yue",
    booktitle = "Proceedings of the 2022 Conference on Empirical Methods in Natural Language Processing",
    month = dec,
    year = "2022",
    address = "Abu Dhabi, United Arab Emirates",
    publisher = "Association for Computational Linguistics",
    url = "https://aclanthology.org/2022.emnlp-main.376/",
    doi = "10.18653/v1/2022.emnlp-main.376",
    pages = "5571--5587",
}

@inproceedings{ma-etal-2023-dice,
    title = "{DICE}: Data-Efficient Clinical Event Extraction with Generative Models",
    author = "Ma, Mingyu Derek  and
      Taylor, Alexander  and
      Wang, Wei  and
      Peng, Nanyun",
    editor = "Rogers, Anna  and
      Boyd-Graber, Jordan  and
      Okazaki, Naoaki",
    booktitle = "Proceedings of the 61st Annual Meeting of the Association for Computational Linguistics (Volume 1: Long Papers)",
    month = jul,
    year = "2023",
    address = "Toronto, Canada",
    publisher = "Association for Computational Linguistics",
    url = "https://aclanthology.org/2023.acl-long.886/",
    doi = "10.18653/v1/2023.acl-long.886",
    pages = "15898--15917",
}

@inproceedings{du-cardie-2020-event,
    title = "Event Extraction by Answering (Almost) Natural Questions",
    author = "Du, Xinya  and
      Cardie, Claire",
    editor = "Webber, Bonnie  and
      Cohn, Trevor  and
      He, Yulan  and
      Liu, Yang",
    booktitle = "Proceedings of the 2020 Conference on Empirical Methods in Natural Language Processing (EMNLP)",
    month = nov,
    year = "2020",
    address = "Online",
    publisher = "Association for Computational Linguistics",
    url = "https://aclanthology.org/2020.emnlp-main.49/",
    doi = "10.18653/v1/2020.emnlp-main.49",
    pages = "671--683",
}

@inproceedings{lu-etal-2023-event,
    title = "Event Extraction as Question Generation and Answering",
    author = "Lu, Di  and
      Ran, Shihao  and
      Tetreault, Joel  and
      Jaimes, Alejandro",
    editor = "Rogers, Anna  and
      Boyd-Graber, Jordan  and
      Okazaki, Naoaki",
    booktitle = "Proceedings of the 61st Annual Meeting of the Association for Computational Linguistics (Volume 2: Short Papers)",
    month = jul,
    year = "2023",
    address = "Toronto, Canada",
    publisher = "Association for Computational Linguistics",
    url = "https://aclanthology.org/2023.acl-short.143/",
    doi = "10.18653/v1/2023.acl-short.143",
    pages = "1666--1688",
}

@inproceedings{uddin-etal-2024-generating,
    title = "Generating Uncontextualized and Contextualized Questions for Document-Level Event Argument Extraction",
    author = "Uddin, Md Nayem  and
      George, Enfa Rose  and
      Blanco, Eduardo  and
      Corman, Steven",
    editor = "Duh, Kevin  and
      Gomez, Helena  and
      Bethard, Steven",
    booktitle = "Proceedings of the 2024 Conference of the North American Chapter of the Association for Computational Linguistics: Human Language Technologies (Volume 1: Long Papers)",
    month = jun,
    year = "2024",
    address = "Mexico City, Mexico",
    publisher = "Association for Computational Linguistics",
    url = "https://aclanthology.org/2024.naacl-long.312/",
    doi = "10.18653/v1/2024.naacl-long.312",
    pages = "5612--5627",
}

@inproceedings{hsu-etal-2022-degree,
    title = "{DEGREE}: A Data-Efficient Generation-Based Event Extraction Model",
    author = "Hsu, I-Hung  and
      Huang, Kuan-Hao  and
      Boschee, Elizabeth  and
      Miller, Scott  and
      Natarajan, Prem  and
      Chang, Kai-Wei  and
      Peng, Nanyun",
    editor = "Carpuat, Marine  and
      de Marneffe, Marie-Catherine  and
      Meza Ruiz, Ivan Vladimir",
    booktitle = "Proceedings of the 2022 Conference of the North American Chapter of the Association for Computational Linguistics: Human Language Technologies",
    month = jul,
    year = "2022",
    address = "Seattle, United States",
    publisher = "Association for Computational Linguistics",
    url = "https://aclanthology.org/2022.naacl-main.138/",
    doi = "10.18653/v1/2022.naacl-main.138",
    pages = "1890--1908",
}

@inproceedings{hong-liu-2024-towards,
    title = "Towards Better Question Generation in {QA}-based Event Extraction",
    author = "Hong, Zijin  and
      Liu, Jian",
    editor = "Ku, Lun-Wei  and
      Martins, Andre  and
      Srikumar, Vivek",
    booktitle = "Findings of the Association for Computational Linguistics: ACL 2024",
    month = aug,
    year = "2024",
    address = "Bangkok, Thailand",
    publisher = "Association for Computational Linguistics",
    url = "https://aclanthology.org/2024.findings-acl.535/",
    doi = "10.18653/v1/2024.findings-acl.535",
    pages = "9025--9038",
}

@inproceedings{fane-etal-2025-bemeae,
    title = "{BEMEAE}: Moving Beyond Exact Span Match for Event Argument Extraction",
    author = "Fane, Enfa  and
      Uddin, Md Nayem  and
      Ikumariegbe, Oghenevovwe  and
      Kashif, Daniyal  and
      Blanco, Eduardo  and
      Corman, Steven",
    editor = "Chiruzzo, Luis  and
      Ritter, Alan  and
      Wang, Lu",
    booktitle = "Proceedings of the 2025 Conference of the Nations of the Americas Chapter of the Association for Computational Linguistics: Human Language Technologies (Volume 1: Long Papers)",
    month = apr,
    year = "2025",
    address = "Albuquerque, New Mexico",
    publisher = "Association for Computational Linguistics",
    url = "https://aclanthology.org/2025.naacl-long.295/",
    doi = "10.18653/v1/2025.naacl-long.295",
    pages = "5734--5749",
    ISBN = "979-8-89176-189-6",
}

@misc{lu2025exactmatchsemanticallyreassessing,
      title={Beyond Exact Match: Semantically Reassessing Event Extraction by Large Language Models}, 
      author={Yi-Fan Lu and Xian-Ling Mao and Tian Lan and Heyan Huang and Chen Xu and Xiaoyan Gao},
      year={2025},
      eprint={2410.09418},
      archivePrefix={arXiv},
      primaryClass={cs.CL},
      url={https://arxiv.org/abs/2410.09418}, 
}

@article {Caufield19009118,
	author = {Caufield, J. Harry and Zhou, Yichao and Bai, Yunsheng and Liem, David A. and Garlid, Anders O. and Chang, Kai-Wei and Sun, Yizhou and Ping, Peipei and Wang, Wei},
	title = {A Comprehensive Typing System for Information Extraction from Clinical Narratives},
	elocation-id = {19009118},
	year = {2019},
	doi = {10.1101/19009118},
	publisher = {Cold Spring Harbor Laboratory Press},
	URL = {https://www.medrxiv.org/content/early/2019/10/22/19009118},
	eprint = {https://www.medrxiv.org/content/early/2019/10/22/19009118.full.pdf},
	journal = {medRxiv}
}

@misc{qwen3technicalreport,
      title={Qwen3 Technical Report}, 
      author={Qwen Team},
      year={2025},
      eprint={2505.09388},
      archivePrefix={arXiv},
      primaryClass={cs.CL},
      url={https://arxiv.org/abs/2505.09388}, 
}

@inproceedings{reimers-2019-sentence-bert,
  title = "Sentence-BERT: Sentence Embeddings using Siamese BERT-Networks",
  author = "Reimers, Nils and Gurevych, Iryna",
  booktitle = "Proceedings of the 2019 Conference on Empirical Methods in Natural Language Processing",
  month = "11",
  year = "2019",
  publisher = "Association for Computational Linguistics",
  url = "https://arxiv.org/abs/1908.10084",
}

@inproceedings{liu-etal-2020-event,
    title = "Event Extraction as Machine Reading Comprehension",
    author = "Liu, Jian  and
      Chen, Yubo  and
      Liu, Kang  and
      Bi, Wei  and
      Liu, Xiaojiang",
    editor = "Webber, Bonnie  and
      Cohn, Trevor  and
      He, Yulan  and
      Liu, Yang",
    booktitle = "Proceedings of the 2020 Conference on Empirical Methods in Natural Language Processing (EMNLP)",
    month = nov,
    year = "2020",
    address = "Online",
    publisher = "Association for Computational Linguistics",
    url = "https://aclanthology.org/2020.emnlp-main.128/",
    doi = "10.18653/v1/2020.emnlp-main.128",
    pages = "1641--1651",
}

@inproceedings{li-etal-2020-event,
    title = "Event Extraction as Multi-turn Question Answering",
    author = "Li, Fayuan  and
      Peng, Weihua  and
      Chen, Yuguang  and
      Wang, Quan  and
      Pan, Lu  and
      Lyu, Yajuan  and
      Zhu, Yong",
    editor = "Cohn, Trevor  and
      He, Yulan  and
      Liu, Yang",
    booktitle = "Findings of the Association for Computational Linguistics: EMNLP 2020",
    month = nov,
    year = "2020",
    address = "Online",
    publisher = "Association for Computational Linguistics",
    url = "https://aclanthology.org/2020.findings-emnlp.73/",
    doi = "10.18653/v1/2020.findings-emnlp.73",
    pages = "829--838",
}

@inproceedings{huang-etal-2024-textee,
    title = "{T}ext{EE}: Benchmark, Reevaluation, Reflections, and Future Challenges in Event Extraction",
    author = "Huang, Kuan-Hao  and
      Hsu, I-Hung  and
      Parekh, Tanmay  and
      Xie, Zhiyu  and
      Zhang, Zixuan  and
      Natarajan, Prem  and
      Chang, Kai-Wei  and
      Peng, Nanyun  and
      Ji, Heng",
    editor = "Ku, Lun-Wei  and
      Martins, Andre  and
      Srikumar, Vivek",
    booktitle = "Findings of the Association for Computational Linguistics: ACL 2024",
    month = aug,
    year = "2024",
    address = "Bangkok, Thailand",
    publisher = "Association for Computational Linguistics",
    url = "https://aclanthology.org/2024.findings-acl.760/",
    doi = "10.18653/v1/2024.findings-acl.760",
    pages = "12804--12825",
}

@inproceedings{srivastava-etal-2025-instruction,
    title = "Instruction-Tuning {LLM}s for Event Extraction with Annotation Guidelines",
    author = "Srivastava, Saurabh  and
      Pati, Sweta  and
      Yao, Ziyu",
    editor = "Che, Wanxiang  and
      Nabende, Joyce  and
      Shutova, Ekaterina  and
      Pilehvar, Mohammad Taher",
    booktitle = "Findings of the Association for Computational Linguistics: ACL 2025",
    month = jul,
    year = "2025",
    address = "Vienna, Austria",
    publisher = "Association for Computational Linguistics",
    url = "https://aclanthology.org/2025.findings-acl.677/",
    doi = "10.18653/v1/2025.findings-acl.677",
    pages = "13055--13071",
    ISBN = "979-8-89176-256-5",
}

@inproceedings{tanev-etal-2025-exploring,
    title = "Exploring the Performance of Large Language Models for Event Detection and Extraction in the Health Domain",
    author = "Tanev, Hristo  and
      Stefanovitch, Nicolas  and
      Harmatha, Tom{\'a}{\v{s}}  and
      F. Sousa, Diana",
    editor = "Angelova, Galia  and
      Kunilovskaya, Maria  and
      Escribe, Marie  and
      Mitkov, Ruslan",
    booktitle = "Proceedings of the 15th International Conference on Recent Advances in Natural Language Processing - Natural Language Processing in the Generative AI Era",
    month = sep,
    year = "2025",
    address = "Varna, Bulgaria",
    publisher = "INCOMA Ltd., Shoumen, Bulgaria",
    url = "https://aclanthology.org/2025.ranlp-1.143/",
    pages = "1237--1247",
}

@inproceedings{shimizu-etal-2025-exploring,
    title = "Exploring {LLM} Annotation for Adaptation of Clinical Information Extraction Models under Data-sharing Restrictions",
    author = "Shimizu, Seiji  and
      Shohei, Hisada  and
      Uno, Yutaka  and
      Yada, Shuntaro  and
      Wakamiya, Shoko  and
      Aramaki, Eiji",
    editor = "Che, Wanxiang  and
      Nabende, Joyce  and
      Shutova, Ekaterina  and
      Pilehvar, Mohammad Taher",
    booktitle = "Findings of the Association for Computational Linguistics: ACL 2025",
    month = jul,
    year = "2025",
    address = "Vienna, Austria",
    publisher = "Association for Computational Linguistics",
    url = "https://aclanthology.org/2025.findings-acl.757/",
    doi = "10.18653/v1/2025.findings-acl.757",
    pages = "14678--14694",
    ISBN = "979-8-89176-256-5",
}

@article{10.1093/jamiaopen/ooaf109,
    author = {Builtjes, Luc and Bosma, Joeran and Prokop, Mathias and van Ginneken, Bram and Hering, Alessa},
    title = {Leveraging open-source large language models for clinical information extraction in resource-constrained settings},
    journal = {JAMIA Open},
    volume = {8},
    number = {5},
    pages = {ooaf109},
    year = {2025},
    month = {10},
    issn = {2574-2531},
    doi = {10.1093/jamiaopen/ooaf109},
    url = {https://doi.org/10.1093/jamiaopen/ooaf109},
    eprint = {https://academic.oup.com/jamiaopen/article-pdf/8/5/ooaf109/64465454/ooaf109.pdf},
}

@inproceedings{zhang-etal-2024-ultra,
    title = "{ULTRA}: Unleash {LLM}s' Potential for Event Argument Extraction through Hierarchical Modeling and Pair-wise Self-Refinement",
    author = "Zhang, Xinliang Frederick  and
      Blum, Carter  and
      Choji, Temma  and
      Shah, Shalin  and
      Vempala, Alakananda",
    editor = "Ku, Lun-Wei  and
      Martins, Andre  and
      Srikumar, Vivek",
    booktitle = "Findings of the Association for Computational Linguistics: ACL 2024",
    month = aug,
    year = "2024",
    address = "Bangkok, Thailand",
    publisher = "Association for Computational Linguistics",
    url = "https://aclanthology.org/2024.findings-acl.487/",
    doi = "10.18653/v1/2024.findings-acl.487",
    pages = "8172--8185",
}

@inproceedings{uddin-etal-2024-asking,
    title = "Asking and Answering Questions to Extract Event-Argument Structures",
    author = "Uddin, Md Nayem  and
      George, Enfa Rose  and
      Blanco, Eduardo  and
      Corman, Steven R.",
    editor = "Calzolari, Nicoletta  and
      Kan, Min-Yen  and
      Hoste, Veronique  and
      Lenci, Alessandro  and
      Sakti, Sakriani  and
      Xue, Nianwen",
    booktitle = "Proceedings of the 2024 Joint International Conference on Computational Linguistics, Language Resources and Evaluation (LREC-COLING 2024)",
    month = may,
    year = "2024",
    address = "Torino, Italia",
    publisher = "ELRA and ICCL",
    url = "https://aclanthology.org/2024.lrec-main.142/",
    pages = "1609--1626",
}

@inproceedings{tong-etal-2022-docee,
    title = "{D}oc{EE}: A Large-Scale and Fine-grained Benchmark for Document-level Event Extraction",
    author = "Tong, MeiHan  and
      Xu, Bin  and
      Wang, Shuai  and
      Han, Meihuan  and
      Cao, Yixin  and
      Zhu, Jiangqi  and
      Chen, Siyu  and
      Hou, Lei  and
      Li, Juanzi",
    editor = "Carpuat, Marine  and
      de Marneffe, Marie-Catherine  and
      Meza Ruiz, Ivan Vladimir",
    booktitle = "Proceedings of the 2022 Conference of the North American Chapter of the Association for Computational Linguistics: Human Language Technologies",
    month = jul,
    year = "2022",
    address = "Seattle, United States",
    publisher = "Association for Computational Linguistics",
    url = "https://aclanthology.org/2022.naacl-main.291/",
    doi = "10.18653/v1/2022.naacl-main.291",
    pages = "3970--3982",
}

@inproceedings{parekh-etal-2023-geneva,
    title = "{GENEVA}: Benchmarking Generalizability for Event Argument Extraction with Hundreds of Event Types and Argument Roles",
    author = "Parekh, Tanmay  and
      Hsu, I-Hung  and
      Huang, Kuan-Hao  and
      Chang, Kai-Wei  and
      Peng, Nanyun",
    editor = "Rogers, Anna  and
      Boyd-Graber, Jordan  and
      Okazaki, Naoaki",
    booktitle = "Proceedings of the 61st Annual Meeting of the Association for Computational Linguistics (Volume 1: Long Papers)",
    month = jul,
    year = "2023",
    address = "Toronto, Canada",
    publisher = "Association for Computational Linguistics",
    url = "https://aclanthology.org/2023.acl-long.203/",
    doi = "10.18653/v1/2023.acl-long.203",
    pages = "3664--3686",
}

@inproceedings{muc-1992-message,
    title = "Overview of the Fourth {M}essage {U}nderstanding {E}valuation and {C}onference",
    author = "Sundheim, Beth M.",
    booktitle = "{F}ourth {M}essage {U}nderstanding {C}onference ({MUC}-4): Proceedings of a Conference Held in {M}c{L}ean, {V}irginia, {J}une 16-18, 1992",
    year = "1992",
    url = "https://aclanthology.org/M92-1001/"
}

@inproceedings{pryzant2023automatic,
title = "Automatic Prompt Optimization with ``Gradient Descent'' and Beam Search",
    author = "Pryzant, Reid  and
      Iter, Dan  and
      Li, Jerry  and
      Lee, Yin  and
      Zhu, Chenguang  and
      Zeng, Michael",
    editor = "Bouamor, Houda  and
      Pino, Juan  and
      Bali, Kalika",
    booktitle = "Proceedings of the 2023 Conference on Empirical Methods in Natural Language Processing",
    month = dec,
    year = "2023",
    address = "Singapore",
    publisher = "Association for Computational Linguistics",
    url = "https://aclanthology.org/2023.emnlp-main.494/",
    doi = "10.18653/v1/2023.emnlp-main.494",
    pages = "7957--7968",

}

@article{yuksekgonul2025optimizing,
  title={Optimizing generative AI by backpropagating language model feedback},
  author={Yuksekgonul, Mert and Bianchi, Federico and Boen, Joseph and Liu, Sheng and Lu, Pan and Huang, Zhi and Guestrin, Carlos and Zou, James},
  journal={Nature},
  volume={639},
  number={8055},
  pages={609--616},
  year={2025},
  publisher={Nature Publishing Group}
}

@article{lee2025feedback,
  title={Feedback descent: Open-ended text optimization via pairwise comparison},
  author={Lee, Yoonho and Boen, Joseph and Finn, Chelsea},
  journal={arXiv preprint arXiv:2511.07919},
  year={2025}
}

@inproceedings{cherep2026visualpersuasion,
  title={Visual Persuasion: What Influences Decisions of Vision-Language Models?},
  author={Cherep, Manuel and M R, Pranav and Maes, Pattie and Singh, Nikhil},
  year={2026},
  url={https://arxiv.org/abs/2602.15278}
}

\appendix

\begin{figure*}[h!]
  \centering
  \includegraphics[width =\linewidth]{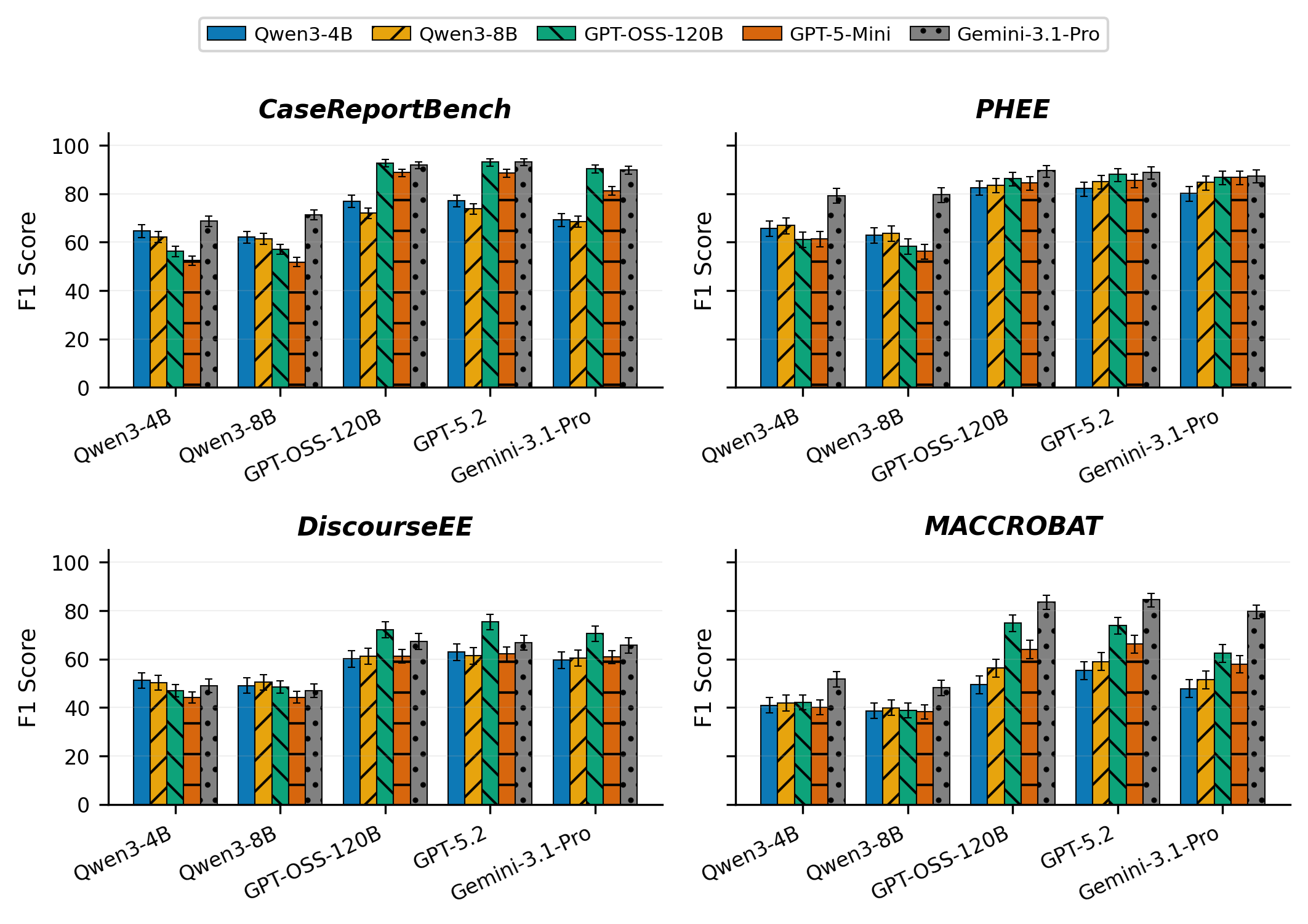}
\caption{Per-model breakdown of zero-shot question
generation quality. Each group of bars shows F1 scores
for a single \QG~model evaluated with all five prediction
models on each dataset. Larger \QG~models (GPT-OSS-120B, GPT-5.2, Gemini-3.1-Pro) consistently outperforms the
smaller Qwen variants across all datasets and prediction
models. Averages are reported in
Table~\ref{table:qg_ablation}.}
\label{figure:qg-ablation}
\end{figure*}

\begin{table*}[!htb]
\centering
\small
\renewcommand*{\arraystretch}{1}
\begin{tabular}{l|C{1cm}C{1cm}C{0.8cm}C{0.8cm}C{1cm}|C{1cm}C{1cm}C{0.8cm}C{0.8cm}C{1cm}}
  \rowcolor{cyan!10} & Qwen3 4B & Qwen3 8B & OSS 120B & Gpt-5 mini & Gemini 3.1-pro & Qwen3 4B & Qwen3 8B & OSS 120B & Gpt-5 mini & Gemini 3.1-pro \\
\midrule
\textbf{QG Model} & \multicolumn{5}{c|}{\textit{\textbf{CaseReportBench}}} & \multicolumn{5}{c}{\textit{\textbf{PHEE}}} \\
\midrule
Qwen3-4B$^{\text{ft-b}}$ & 60.64 & 59.43 & 60.68 & 62.17 & 60.95 & 75.76 & 78.60 & 79.89 & 79.69 & 82.20 \\
Qwen3-4B$^{\text{ft-a}}$ & \textbf{62.56} & \textbf{64.09} & \textbf{71.63} & \textbf{67.86} & \textbf{71.32} & 78.22 & \textbf{81.52} & 81.00 & 78.97 & 83.92 \\
Qwen3-4B$^{\text{ft-in}}$ & 58.98 & 60.01 & 63.15 & 61.59 & 60.77 & \textbf{78.48} & 80.24 & \textbf{83.61} & \textbf{80.30} & \textbf{84.96} \\
\multicolumn{11}{c}{- - - - - - - - - - - - - - - - - - - - - - - - - - - - - - - - - - - - - - - - - - - - - - - - - - - - - - - - - - - - - - - - - - - - - - - - - - - - - - - - - - } \\
Qwen3-8B$^{\text{ft-b}}$ & 62.98 & 61.59 & 69.40 & 66.07 & 67.89 & 77.42 & 77.38 & 78.88 & 77.18 & 81.35 \\
Qwen3-8B$^{\text{ft-a}}$ & 60.89 & 61.03 & 69.22 & 65.64 & 66.83 & \textbf{78.07} & \textbf{80.66} & 82.36 & \textbf{82.74} & \textbf{84.25} \\
Qwen3-8B$^{\text{ft-in}}$ & \textbf{63.85} & \textbf{64.59} & \textbf{70.88} & \textbf{69.16} & \textbf{69.98} & 77.86 & 79.99 & \textbf{84.52} & 80.03 & 83.70 \\
\midrule
\midrule
 & \multicolumn{5}{c|}{\textit{\textbf{DiscourseEE}}} & \multicolumn{5}{c}{\textit{\textbf{MACCROBAT}}} \\
\midrule
Qwen3-4B$^{\text{ft-b}}$ & 45.27 & 47.61 & 49.36 & 50.38 & 46.22 & 40.61 & 45.22 & 49.52 & 45.45 & 53.18 \\
Qwen3-4B$^{\text{ft-a}}$ & \textbf{49.75} & \textbf{50.16} & \textbf{54.12} & \textbf{54.01} & \textbf{51.46} & 41.71 & \textbf{46.37} & \textbf{51.39} & \textbf{48.62} & \textbf{56.84} \\
Qwen3-4B$^{\text{ft-in}}$ & 47.69 & 48.57 & 51.35 & 51.88 & 45.96 & \textbf{45.37} & 45.97 & 49.20 & 46.54 & 55.34 \\
\multicolumn{11}{c}{- - - - - - - - - - - - - - - - - - - - - - - - - - - - - - - - - - - - - - - - - - - - - - - - - - - - - - - - - - - - - - - - - - - - - - - - - - - - - - - - - - } \\
Qwen3-8B$^{\text{ft-b}}$ & 47.11 & 49.20 & \textbf{56.74} & \textbf{54.47} & \textbf{52.71} & \textbf{41.81} & \textbf{44.96} & 50.64 & 46.24 & 56.22 \\
Qwen3-8B$^{\text{ft-a}}$ & 47.72 & \textbf{50.57} & 55.66 & 53.21 & 50.21 & 41.02 & 43.25 & \textbf{50.96} & \textbf{47.53} & 53.36 \\
Qwen3-8B$^{\text{ft-in}}$ & \textbf{47.93} & 47.63 & 50.86 & 53.37 & 48.73 & 40.95 & 44.69 & 50.90 & 45.45 & \textbf{56.74} \\
\bottomrule
\end{tabular}
\caption{Impact of fine-tuning data composition on QG model performance. Each row fine-tunes the QG model (Qwen3-4B or Qwen3-8B) on a different subset of \FeedQ-optimized questions: \textit{ft-b} uses a balanced mix sampling equally across datasets, \textit{ft-a} uses all available optimized questions regardless of source distribution, and \textit{ft-in} uses only in-domain questions matching the target dataset. The best F1 score of the ablation is reported in Table \ref{table:test_results_cm}.}
\label{table:ft_ablation}
\end{table*}


\newpage
\null
\newpage
\section{Datasets}
\label{apx:dataset-details}
All datasets are converted to the unified format described
in Section~\ref{sec:preliminaries}: each sample consists of
a document, a role, and its ground-truth arguments. We
create development and test splits of 500 samples per
dataset, except CaseReportBench where the smaller original
dataset limits both to 350.
Table~\ref{table:dataset-statistics} reports detailed
statistics. Descriptions of each dataset provided in the following. 

\begin{table*}[!htb]
\small
\renewcommand*{\arraystretch}{1}
\centering
\begin{tabular}{l|ccc|c|cC{1.5cm}C{2cm}|c}
\toprule
\textbf{Dataset} & \multicolumn{4}{c|}{\textbf{Data Splits}} & \multicolumn{4}{c}{\textbf{General Statistics}} \\
\midrule
 & Train & Dev & Test & Total & \#Roles & Avg. doc length & Avg. argument length  &  Domain\\
\midrule
CaseReportBench & 620  & 350 & 350 & 1320 & 18 & 581.7 & 20.71 & Clinical Reports \\
PHEE            & 5000 & 500 & 500 & 6000 & 14 & 19.7  & 2.45  & Pharmacovigilance texts \\
DiscourseEE     & 2200 & 500 & 500 & 3200 & 34 & 120.8 & 3.04  & Online Health Forums \\
MACCROBAT       & 5000 & 500 & 500 & 6000 & 22 & 24.2  & 1.73  & PubMed Reports \\
\bottomrule
\end{tabular}
\caption{Summary statistics of the four datasets used in our experiments, including split sizes and general statistics.}
\label{table:dataset-statistics}
\end{table*}

\paragraph{CaseReportBench} \cite{pmlr-v287-zhang25b}. is a dense information extraction dataset from clinical case reports on rare diseases. Each case report is annotated by clinical experts into different categories such as laboratory findings, diagnosis, treatment, and outcomes. This dataset targets the extraction of structured information from long, unstructured clinical narratives requiring reasoning over entire documents. We source the dataset from author's HuggigFace repository \href{https://huggingface.co/datasets/cxyzhang/caseReportBench_ClinicalDenseExtraction_Benchmark}{https://huggingface.com/CaseReportBench}. We map it into our unified format by treating each category as a role with its corresponding argument strings.

\paragraph{PHEE} \cite{sun-etal-2022-phee}. is an argument extraction dataset sourced for the pharmacovigilance (i.e. drug safety) domain. The dataset contains 14 different roles covering aspects such as drug, dosage, route, frequency, duration, disorder, and patient demographics (age, gender, race). We obtain the dataset from official repo \href{https://github.com/ZhaoyueSun/PHEE}{https://github.com/ZhaoyueSun/PHEE}. The original dataset includes both trigger and argument span annotations at the sentence level. Following our formulation, we only take the argument strings and combine multiple arguments under the same role into a single argument list. We select 6000 randomly selected samples out of 22,274 document-role samples. 

\paragraph{DiscourseEE} \cite{sharif-etal-2024-explicit}. A dataset annotated from online health forum about information seeking posts on opioid use disorder. It features 34 unique roles across with all arguments annotated as free-text strings. Notably, 51.2\% of the arguments are implicit (inferable from context but not directly stated) making this dataset uniquely challenging for argument extraction. We sourced the data from official repository \url{https://github.com/omar-sharif03/DiscourseEE/tree/main/Data} and transform it into our format following author’s provided code.

\paragraph{MACCROBAT} \cite{ma-etal-2023-dice}. is a clinical event extraction dataset repurposed from expert-annotated PubMed case reports originally introduced by \citet{Caufield19009118}. It contains 22 argument roles inferred from entity-event modification relations. The dataset is notably dense, averaging 3.21 events per sentence. We collect the repurposed dataset from official GitHub repository \href{https://github.com/mdkma/DICE/tree/main/maccrobat/Data}{https://github.com/DICE}. We select 6000 randomly selected samples out of 6,882 document-role samples.

\section{Implementation Details}
\label{apx:implementaion-details}
\subsection{Models}
\label{apdx:models}
We evaluate across three model families \textit{Qwen}, \textit{GPT}, and \textit{Gemini} — spanning open-source and proprietary LLMs at different scales. For the Qwen \cite{qwen3technicalreport}, we use the Qwen3-4B and Qwen3-8B in non-thinking mode. Beyond prediction, we fine-tune both models with LoRA for question generation (Section~\ref{sec:question-generation}). We use three GPT\footnote{\url{https://openai.com/index/introducing-gpt-5-2/}}  models: Gpt-OSS-120B serves as both a prediction and question generation model with identical hyperparameters; GPT-5-mini with medium reasoning effort serves as a prediction model; and GPT-5.2 with high reasoning effort serves as a strong question generation model. Finally, we evaluate Gemini-3.1-Pro\footnote{\url{https://deepmind.google/models/gemini/pro/}} with high reasoning for both question generation and prediction. 

\subsection{\FeedQ~Parameters and Leakage Analysis}
\label{apdx:leakage-analysis}
We run \FeedQ~for up to 5 iterations with a patience of 3 and a target
score of 1.0 (i.e., full extraction of all ground-truth arguments).
Figure~\ref{fig:iter_dist} confirms that 5 iterations are sufficient:
across all four datasets, the best-scoring question set is found within
this budget, though the number of iterations required varies by dataset
complexity. 

\paragraph{Leakage check.}
Because \FeedQ~optimizes questions against ground-truth arguments, the optimizer could embed answer fragments in the question text. We guard against this with two measures. First, the question refinement prompt (Appendix~\ref{prompt:question-refinement}) explicitly instructs the model not to include answer-related information. Second, a standalone leakage checker inspects every generated question set and removes any question containing ground-truth content. Here, we analyze the checker's behavior systematically on the development set. Table~\ref{table:leakage} reports the proportion of iterations in which the leakage checker flags at least one generated question as containing ground-truth content \textit{before} filtering. Leakage occurs substantially across all four datasets, affecting 42--55\% of iterations and 45--65\% of samples. \textbf{The checker removes these questions before they enter the QG training set}, preventing downstream models learning from a flawed signal. 

\updt{To validate the checker, we manually inspect 600 filtered question sets (150 per dataset) for missed leakage. \textbf{The false negative rate is 0.83\% (5/600), confirming that the filter reliably catches answer-leaking questions}. Beyond filtering, our test-time setup eliminates ground-truth access entirely: fine-tuned QG models generate questions for unseen queries with no oracle, and outperform strong QG models and all knowledge-driven baselines on average (Tables~\ref{table:test_results_cm}--\ref{table:test_avg_cm}). If the gains were driven by implicit encoding of target information during \FeedQ~optimization, we would expect this transfer to degrade. Instead, it improves over the base model. The gain also holds under domain shift: on DocEE (out-of-domain), fine-tuning improves Qwen3-8B by +7.4 F1 (Table~\ref{table:ood_results_avg}). Implicit answer encoding tied to the optimization documents or semantic leakage would not transfer to a new domain with different argument distributions. Taken together, these analyses indicate minimal leakage.}

\begin{table}[!htb]
\small
\centering
\begin{tabular}{lcc}
\toprule
\textbf{Dataset} & \textbf{Iters Leaked (\%)} & \textbf{Samples w/ Leak (\%)} \\
\midrule
CRB & 280 / 505\phantom{0} (55.4) & 219 / 350 (62.6) \\
PHEE            & 272 / 644\phantom{0} (42.2) & 223 / 500 (44.6) \\
DEE     & 516 / 1104 (46.7) & 323 / 500 (64.6) \\
MBAT       & 353 / 816\phantom{0} (43.3) & 228 / 500 (45.6) \\
\bottomrule
\end{tabular}
\caption{Ground-truth leakage in \FeedQ-generated questions on the dev set, measured before filtering. \textit{Iter.\ Leaked} counts optimization iterations that produced at least one question containing a verbatim argument; \textit{Samples w/ Leak} counts unique samples affected.}
\label{table:leakage}
\end{table}

\subsection{Knowledge Questions Source} 
\label{knowledge-q-source}
 \updt{We take these human-written questions directly from materials released by the dataset authors, or minimally adapt them to QA form from the expert annotation guidelines. For DiscourseEE, we use the expert questions released with the original paper \cite{sharif-etal-2024-explicit} verbatim. For CaseReportBench \cite{pmlr-v287-zhang25b}, we adapt the role definitions in the Expert Annotation Guidelines released in the dataset's GitHub repository.\footnote{\url{https://tinyurl.com/vvrautsr}} For PHEE \cite{sun-etal-2022-phee} and MACCROBAT \cite{ma-etal-2023-dice,Caufield19009118}, we adapt the role-level definitions provided in the papers. Therefore, this constitutes a strong human baseline questions.}

\subsection{Fine-tuning}
\label{sec:fine-tuning-details}

We fine-tune Qwen3-4B and Qwen3-8B using LoRA (rank 32) with Tinker\footnote{https://tinker-docs.thinkingmachines.ai/supervised-learning}. All models are trained for 2 epochs with a learning rate of $3.7e^{-4}$ and maximum sequence length of 4,096. We construct three training mixes (Table \ref{tab:ft-mixes}): a balanced mix (6,200 triples) that caps each dataset at a comparable count, all-available mix (12,820) triples that uses all the samples, and \textit{in-domain} mix that trains on only the target dataset’s triples. In all cases, 20\% is reserved for validation to select the best checkpoint. We report high score per model in Table \ref{table:test_results_cm} and provide the full breakdown in Table \ref{table:ft_ablation}.

\begin{table}[!htb]
\centering
\small
\begin{tabular}{lcc}
\toprule
Dataset & Balanced (ft-b) & All (ft-a) \\
\midrule
CaseReportBench & 620 & 620 \\
PHEE & 1,860 & 5,000 \\
DiscourseEE & 1,860 & 2,200 \\
MACCROBAT & 1,860 & 5,000 \\
\midrule
Total & 6,200 & 12,820 \\
\bottomrule
\end{tabular}
\caption{Number of $(\mathcal{D}, r, \mathcal{Q}^*)$ triples per dataset in each fine-tuning mix. We reserve 20\% for validation to select the best checkpoint and use the remaining 80\% for training.}
\label{tab:ft-mixes}
\end{table}

\subsection{Evaluation}
\label{apdx:evaluation}
We follow REGen~\cite{sharif-etal-2025-regen}, a three-level hierarchical evaluation framework for generative argument matching. A predicted argument is considered correct if it matches a ground-truth argument at any of three levels: exact, similarity-based relaxed, or LLM-as-judge based complex match. Precision/recall/F1 are then measured the standard way. At relaxed match level, two arguments are considered similar if their SBERT embedding similarity \cite{reimers-2019-sentence-bert} exceeds 0.85. For complex matching, we use GPT-OSS-120B as a judge model with the prompt below.

\begin{tcolorbox}[
    colback=gray!5,
    colframe=teal!80,
    title=\textbf{Prompt for judge-based matching},
    fonttitle=\bfseries,
    sharp corners,
    boxrule=0.8pt
]
Determine if two arguments refer to the same core entity, value, or claim for the given role.
Match if they convey the same core meaning, even with minor differences in detail, formatting, or specificity.

\medskip
\texttt{Context:} \{context\}\\
\texttt{Role:} \{role\}\\
\texttt{Argument 1:} \{x\}\\
\texttt{Argument 2:} \{y\}\\

\texttt{Answer "yes" or "no" only.}
\end{tcolorbox}

\begin{table*}[!htb]
\centering
\scriptsize
\renewcommand*{\arraystretch}{1}
\begin{tabular}{l|C{0.7cm}C{0.7cm}C{0.8cm}C{0.8cm}C{0.8cm}c|C{0.7cm}C{0.7cm}C{0.8cm}C{0.8cm}C{0.8cm}c}
  \rowcolor{cyan!10} & Qwen3 4B & Qwen3 8B & OSS 120B & Gpt-5 mini & Gemini 3.1-pro & Mean & Qwen3 4B & Qwen3 8B & OSS 120B & Gpt-5 mini & Gemini 3.1-pro & Mean \\
\midrule
\textbf{QG Model} & \multicolumn{6}{c|}{\textit{\textbf{DocEE}}} & \multicolumn{6}{c}{\textit{\textbf{GENEVA}}} \\
\midrule
Qwen3-4B & 56.94 & 48.23 & 36.00 & 36.05 & 55.41 & 46.53 & \textbf{55.37} & \textbf{54.55} & \textbf{54.36} & \textbf{54.57} & 76.79 & \textbf{59.13} \\
Qwen3-8B & 58.26 & 50.44 & 41.22 & 40.82 & 55.28 & 49.20 & 54.07 & 51.26 & 53.93 & 53.46 & \textbf{78.76} & 58.30 \\
\multicolumn{13}{c}{- - - - - - - - - - - - - - - - - - - - - - - - - - - - - - - - - - - - - - - - - - - - - - - - - - - - - - - - - - - - - - - - - - - - - - - - - - - - - - - - - - - - - - - - - - - - - - - - - - - - - - - - - - - -} \\
Qwen3-4B$^{\text{ft-b}}$ & \textbf{59.42} & 51.29 & 51.92 & 55.38 & 53.90 & 54.38 & 48.79 & 47.30 & 47.30 & 49.52 & 63.43 & 51.27 \\
Qwen3-4B$^{\text{ft-a}}$ & 57.98 & 51.63 & 55.01 & 57.80 & 53.97 & 55.28 & 48.18 & 52.21 & 48.87 & 52.86 & 61.96 & 52.82 \\
Qwen3-8B$^{\text{ft-b}}$ & 59.27 & \textbf{52.50} & \textbf{56.22} & \textbf{58.63} & \textbf{56.61} & \textbf{56.65} & 50.59 & 49.56 & 47.47 & 50.44 & 66.35 & 52.88 \\
Qwen3-8B$^{\text{ft-a}}$ & 54.04 & 50.04 & 51.96 & 56.63 & 55.46 & 53.63 & 49.11 & 49.51 & 47.01 & 50.21 & 64.18 & 52.00 \\
\midrule
\midrule
\textbf{QG Model} & \multicolumn{6}{c|}{\textit{\textbf{MUC4}}} & \multicolumn{6}{c}{} \\
\midrule
Qwen3-4B & \textbf{64.08} & \textbf{58.64} & 47.08 & 51.08 & 66.36 & 57.45 & - & - & - & - & - & - \\
Qwen3-8B & 62.64 & 58.48 & 51.86 & 53.08 & \textbf{66.73} & 58.56 & - & - & - & - & - & - \\
\multicolumn{13}{c}{- - - - - - - - - - - - - - - - - - - - - - - - - - - - - - - - - - - - - - - - - - - - - - - - - - - - - - - - - - - - - - - - - - - - - - - - - - - - - - - - - - - - - - - - - - - - - - - - - - - - - - - - - - - -} \\ 
Qwen3-4B$^{\text{ft-b}}$ & 56.04 & 55.18 & 52.80 & 61.40 & 61.07 & 57.30 & - & - & - & - & - & - \\
Qwen3-4B$^{\text{ft-a}}$ & 57.46 & 54.74 & 51.74 & 59.02 & 60.71 & 56.73 & - & - & - & - & - & - \\
Qwen3-8B$^{\text{ft-b}}$ & 59.47 & 56.02 & \textbf{57.46} & \textbf{63.35} & 63.22 & \textbf{59.90} & - & - & - & - & - & - \\
Qwen3-8B$^{\text{ft-a}}$ & 57.82 & 53.74 & 53.38 & 60.02 & 61.94 & 57.38 & - & - & - & - & - & - \\
\bottomrule
\end{tabular}
\caption{\updt{Out-of-domain generalization on DocEE, GENEVA, and MUC4. Non-fine-tuned QG models (top) and fine-tuned variants (bottom). Models are fine-tuned with \FeedQ-optimized questions directly. We report f1 score with two variants (\textit{ft-b}: balanced mix, \textit{ft-a}: all available). Bold indicates the best f1 score in each column within a dataset.}}
\label{table:ood_results_full}
\end{table*}

\subsection{Impact of Data Scaling}
\label{dataset-scaling-analysis}
\begin{figure}[h!]
  \centering
  \includegraphics[width =\linewidth]{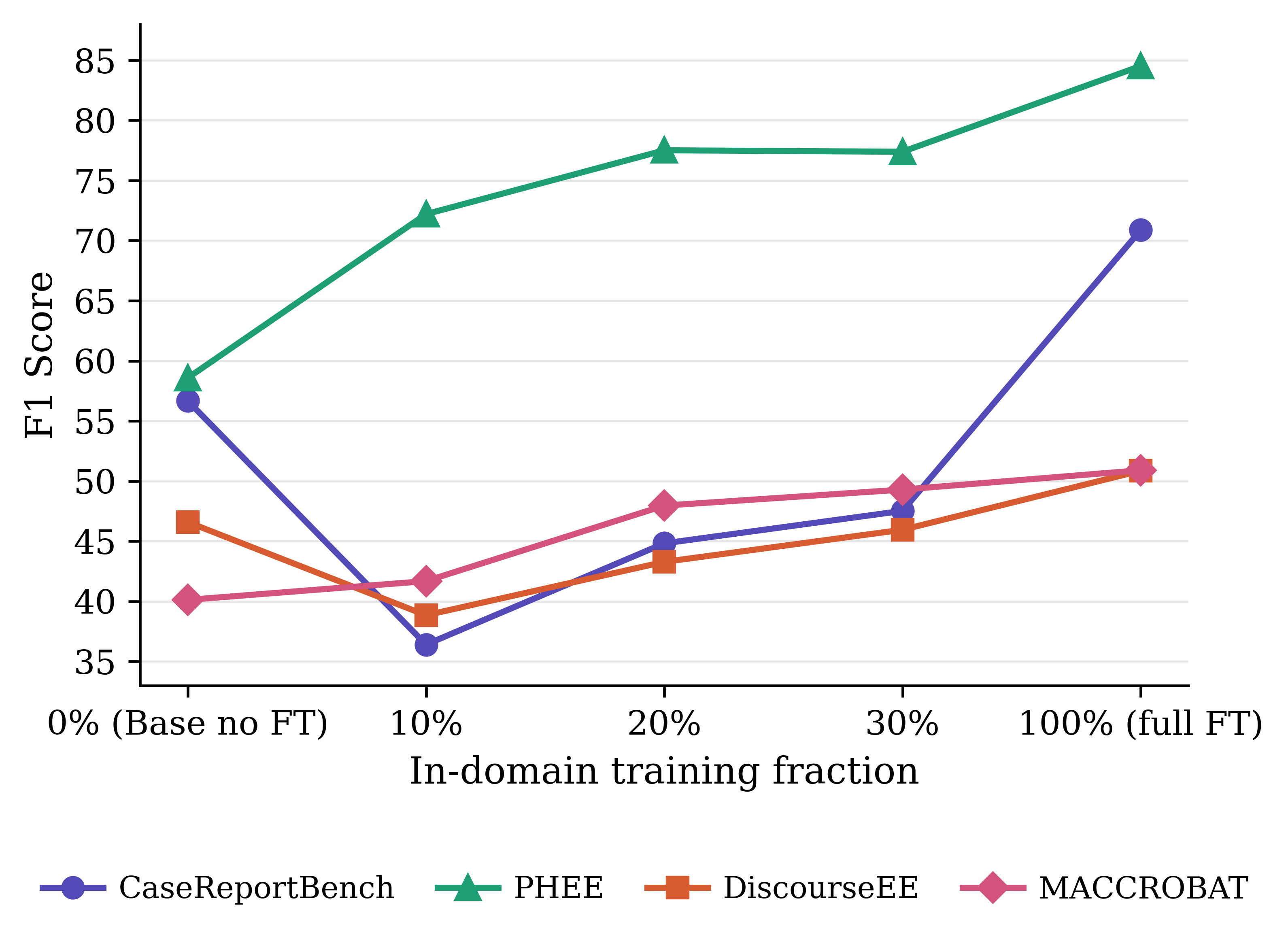}
\caption{Effect of dataset scaling on QG fine-tuning. A separate Qwen3-8B QG model is fine-tuned on 10/20/30/100\% of each dataset's triples.}
\label{fig:data-scaling}
\end{figure}

\updt{We measure how QG fine-tuning scales with training data. We fine-tuned Qwen3-8B QG model on each dataset's randomly sampled triples at 10\%, 20\%, 30\%, and 100\% fractions, and evaluated with GPT-OSS-120B. Figure~\ref{fig:data-scaling} reports the results.}

\updt{PHEE and MACCROBAT are larger, with 5,000 triples each. Even 10\% of the data improves over the base model, and performance rises monotonically as more data is added. CaseReportBench and DiscourseEE behave differently. Both are small: 620 and 2,200 triples in total. At 10\%, both drop below the base model. The drop is consistent with catastrophic forgetting from fine-tuning on a small in-domain sample. Performance recovers as more data is added and surpasses the base by 100\%. Across all four datasets, more training data improves model performance, showing the clear impact of dataset scaling.}

\subsection{Out-of-domain Analysis}
\label{addtional-ood-analysis}
\updt{Fine-tuning improves Qwen3-8B by up to +7.4 F1 on DocEE. DocEE samples are long (often several thousand words), and the base model generates a large pool of loosely relevant questions. The fine-tuned model produces a small set of targeted questions. Fine-tuning generalizes the question-asking behavior beyond the training domains rather than memorizing question patterns specific to them. MUC4 shows the same pattern, with a smaller +1.3 F1 gain. Our approach therefore delivers significant in-domain gains while preserving performance under substantial out-of-domain shift.}

\begin{examplebox1}
\small
\textbf{DocEE example} (full document: 948 words, 5{,}647 chars)\\[2pt]
\textit{Excerpt:} ``The 1931 Hawke's Bay earthquake [\dots] occurred in New Zealand at 10:47 am on 3 February, killing 256, injuring thousands [\dots] Nearly all buildings in the central areas of Napier and Hastings were levelled [\dots] the death toll included 161 people in Napier, 93 in Hastings, and two in Wairoa. Thousands more were injured, with over 400 hospitalised [\dots]''

\textbf{Base model (30 questions, many off-target):}
\begin{itemize}\setlength{\itemsep}{1pt}\setlength{\topsep}{0pt}
  \item How many people were killed in the 1931 Hawke's Bay earthquake?
  \item What was the total number of injuries reported during the earthquake?
  \item [\dots]
\end{itemize}

\textbf{Fine-tuned model (4 questions, targeted to roles and context):}
\begin{itemize}\setlength{\itemsep}{1pt}\setlength{\topsep}{0pt}
  \item What were the total number of deaths and injuries reported from the 1931 Hawke's Bay earthquake?
  \item How many people were killed in each of the affected cities (Napier, Hastings, Wairoa)?
  \item [\dots]
\end{itemize}
\end{examplebox1}

\noindent
\updt{\textbf{Why model fail on GENEVA:} GENEVA's annotation style frequently assigns multiple coreferent or syntactically distinct arguments to the same role within a single sentence. Recovering all such arguments benefits from probing the role from several complementary angles. The broader question sets produced by the base model happen to provide this additional coverage, even though such breadth would introduce noise on datasets with cleaner annotations. The fine-tuned QG, by contrast, produces more focused, role-targeted questions, which improves precision but reduces overall F1 due to lower coverage.}

\begin{examplebox2}
\small
\textbf{GENEVA annotation examples}\\[4pt]

\textbf{Example 1.}~\textit{Excerpt:} ``The best beaches have become home to the finest hotels, which supply almost everything needed for the perfect vacation.''\\[2pt]
\textbf{Role:} \texttt{supply: supplier}\\[1pt]
\textbf{Args:} \{``the finest hotels'', ``which''\}\\[1pt]
\textit{The noun phrase and its relative pronoun are both annotated as gold arguments within the same clause.}\\[6pt]

\textbf{Example 2.}~\textit{Excerpt:} ``Many islands had already transferred their allegiance to the victors, who were led by Philip II of Macedon.''\\[2pt]
\textbf{Role:} \texttt{sending: recipient}\\[1pt]
\textbf{Args:} \{``to the victors'', ``Philip II of Macedon''\}\\[1pt]
\textit{Both the prepositional recipient and the named leader of that group are gold annotations.}\\[6pt]

\textbf{Example 3.}~\textit{Excerpt:} ``\dots defines terrorism as: `The sub-state application of violence or threatened violence\dots' ''\\[2pt]
\textbf{Role:} \texttt{using: agent}\\[1pt]
\textbf{Gold:} \{``sub-state''\}\\[1pt]
\textit{The descriptor is annotated as the agent, requiring a question that targets the modifier rather than the entity.}\\[4pt]
\end{examplebox2}


\begin{table*}[t]
\centering
\small
\setlength{\tabcolsep}{4pt}
\renewcommand{\arraystretch}{1.2}
\begin{tabular}{p{0.5cm} p{4.8cm} p{3.3cm} p{6.2cm}}
\toprule
\textbf{\#} & \textbf{Document Excerpt} & \textbf{Vanilla Questions} & \textbf{List of Questions} \\
\midrule


1 &
\textit{[CaseReportBench, Neuro]} \newline
On general examination, the patients were moderately built with a \textbf{steady gait}. The physical and \textbf{mental development} was also normal. &
What neurological signs, symptoms, or findings are mentioned (e.g., motor skills, reflexes, sensation, mental status)? &
\textbf{Q1.} How was the patient's gait described in the general examination? \newline
\textbf{Q2.} What was reported about the patient's mental development? \\

\midrule

2 &
\textit{[CaseReportBench, Lab/Imaging]} \newline
\textbf{PSA} was 0.393\,ng/mL [\ldots] \textbf{X-ray} of pelvis showed punctate calcification [\ldots] \textbf{CT} scan showed extensive calcified areas [\ldots] \textbf{Biochemical analysis} of urine showed calcium. &
What laboratory test results or imaging findings are reported (e.g., MRI, CT, X-ray, biochemical tests)? &
\textbf{Q1.} Was a prostate specific antigen (PSA) level reported? \newline
\textbf{Q2.} What calcification findings were described on the pelvic X-ray? \newline
\textbf{Q3.} What calcification findings were described on the CT scan? \newline
\textbf{Q4.} What substance was identified in the urine biochemical analysis? \newline
\textbf{Q5.} What components were identified in the biochemical analysis of the extracted prostatic calculi? \\

\midrule


3 &
\textit{[PHEE, Effect]} \newline
A 31-year-old man with OCD developed \textbf{RBD} soon after starting fluoxetine therapy, which \textbf{persisted at PSG study 19 months} after fluoxetine discontinuation. &
What are the outcomes or side effects of the treatments? &
\textbf{Q1.} What sleep disorder did the patient develop after starting fluoxetine therapy? \newline
\textbf{Q2.} According to the polysomnography study, how many months after fluoxetine discontinuation did this disorder persist? \\

\midrule

4 &
\textit{[PHEE, Treatment-Drug]} \newline
We report a case of recurrent torsades de pointes following treatment with \textbf{pentavalent antimonial drugs} and \textbf{amiodarone}. &
Whare the the drugs used as therapy in the event? &
\textbf{Q1.} What antiarrhythmic drug was administered to the patient? \newline
\textbf{Q2.} What antimonial medication was administered to the patient? \\

\midrule


5 &
\textit{[DiscourseEE, Tapering-Type]} \newline
\textbf{Post:} I'm currently seeing an addiction doctor [\ldots] they think withdrawals last only 3 days [\ldots] I'm wondering if a family doctor can prescribe comfort meds? &
Is the subject self-tapering or following prescribed tapering? &
\textbf{Q1.} What instructions did the addiction doctor give about tapering the medication? \newline
\textbf{Q2.} What are the patient's plans for handling the tapering of their medication? \\

\midrule

6 &
\textit{[DiscourseEE, Tapering-Event]} \newline
\textbf{Post:} I took \textbf{6\,mgs} then \textbf{4\,mgs} then \textbf{2\,mg's} yesterday. It was closer to \textbf{2$\frac{1}{2}$ days} [\ldots] I don't plan to take anymore [\ldots] will I have any Suboxone withdrawals? &
What are the tapering steps (drugs, start dosage, duration, goal dosage)? &
\textbf{Q1.} What is the duration of the Suboxone taper before the individual stopped? \newline
\textbf{Q2.} What dosages of Suboxone were taken during the taper and in what order? \newline
\textbf{Q3.} What concerns does the individual express about potential withdrawal? \newline
\textbf{Q4.} What does the individual intend to do regarding Suboxone use after the taper? \\

\midrule


7 &
\textit{[MACCROBAT, Lab\_value]} \newline
Admission blood test showed a \textbf{slight decrease} in hemoglobin (\textbf{106\,g/L}). &
What were the test results, measurements, or laboratory values from the diagnostic procedure? &
\textbf{Q1.} What is the hemoglobin concentration reported in the admission blood test? \newline
\textbf{Q2.} How is the hemoglobin level described qualitatively? \\

\midrule

8 &
\textit{[MACCROBAT, Detailed\_description]} \newline
The patient underwent 6 weeks of \textbf{image-guided} \textbf{external beam} radiation centred on the resection field. &
What are the specific details or characteristics of the therapeutic procedure? &
\textbf{Q1.} What type of radiation therapy was administered? \newline
\textbf{Q2.} What imaging technique was used to guide the radiation treatment? \\

\midrule

9 &
\textit{[MACCROBAT, Dosage]} \newline
She was given [\ldots] sodium bicarbonate (\textbf{8.4\%}, \textbf{10 flacon/day}) infusion for 24 hours before chemotherapy. &
What is the dosage, amount, or quantity of the medication? &
\textbf{Q1.} What is the concentration of the sodium bicarbonate infusion administered? \newline
\textbf{Q2.} What is the daily amount of sodium bicarbonate given? \\

\bottomrule
\end{tabular}
\caption{Examples of LoQ questions compared to the Vanilla-Q counterparts across datasets and roles. Bold text in the document excerpt marks the distinct arguments each optimized question targets. For simple multi-argument roles (rows 7–9), LoQ generates 2 targeted questions; for complex roles it generates a set of questions to extract each argument correctly.}
\label{tab:question-examples}
\end{table*}

\section{Prompts}
\label{section:prompts}
\subsection{Leakage Check}
\begin{promptbox}
\label{prompt:leakage-check}
\small
You are a quality control reviewer for an information extraction
system. Your task is to check whether any of the generated questions
contain \textbf{information leakage}---meaning the questions
directly reveal, embed, or strongly hint at specific ground truth
arguments that they are supposed to help extract.

\medskip
\textbf{Why this matters:} The questions should guide an extraction
model to find relevant information in a document, NOT hand the
answers to it. A leaked question makes extraction trivially easy, but produces questions that won't generalize to new, unseen documents.

\medskip
\textbf{Leakage (MUST be rewritten)}\\
A question has \textbf{strong leakage} if it:\\
1. \textbf{Directly contains} a ground truth argument verbatim or
near-verbatim (e.g., asking ``Is the patient's hemoglobin 7.2
g/dL?'' when ``hemoglobin 7.2 g/dL'' is a ground truth argument)\\
2. \textbf{Embeds specific values, dosage, or measurements} from
the ground truth (e.g., ``Was the patient started on levetiracetam
500mg twice daily?'' when that exact medication regimen is a ground
truth argument)

\medskip
\textbf{Acceptable / Minor Overlap (keep unchanged)}\\
A question is \textbf{acceptable} and should NOT be rewritten if it:\\
- Asks about a general clinical category or organ system (e.g.,
``What hematological abnormalities were identified?'' is fine even
if a specific hematological finding is in the ground truth)\\
- Uses standard medical terminology that naturally overlaps with
ground truth terms (e.g., mentioning ``neurological'' when
neurological findings exist in ground truth)\\
- Asks open-ended questions that probe a relevant clinical domain
without specifying exact answers\\
- References broad clinical concepts like symptom categories,
diagnostic approaches, or treatment classes


\medskip
\textbf{Input}\\
\textbf{Role:} \{role\}\\
\textbf{Current Questions:} \{current\_questions\}\\
\textbf{Ground Truth Arguments:} \{gt\_arguments\}

\medskip
\textbf{Instructions}\\
1. Review each question against all ground truth arguments for
strong leakage only\\
2. For any question with leakage, rewrite it to be more
general while still targeting the same information category\\
3. For all other questions, keep them exactly unchanged

\medskip
Respond ONLY with valid JSON in this exact format:\\
\texttt{\{\{}\\
\texttt{~~"is\_leaked": "yes/no",}\\
\texttt{~~"final\_questions": [}\\
\texttt{~~~~"rewritten or original question 1",}\\
\texttt{~~~~"rewritten or original question 2"}\\
\texttt{~~]}\\
\texttt{\}\}}

\medskip
Set \texttt{"is\_leaked"} to \texttt{"yes"} only if at least one
question had strong leakage and was rewritten. Otherwise, set it to
\texttt{"no"}. Do not include any text outside the JSON.

\end{promptbox}

\subsection{Question Generation}
\begin{promptbox}
\label{prompt:question-generation}
\small
You are a helpful assistant that generates clear and distinct
questions to best extract the ground truth arguments for a given
role.
\medskip

\textbf{Role:} \{role\}\\
\textbf{Document:} \{document\}\\
\textbf{Ground truth arguments:} \{gt\_arguments\}
\medskip

\textbf{Instructions:}\\
-- Each question should aim to extract one or more ground truth
arguments.\\
-- If a single question can logically extract multiple arguments,
group them in one question.\\
-- Avoid redundancy---do not create multiple questions that ask
for the same information.\\
-- Avoid yes/no questions or vague phrasing.\\
-- Keep questions concise so the extraction model does not get
confused and extract irrelevant arguments.
\medskip

\textbf{Examples:}\\
\{``subject'': ``Individual or patient involved in the medical
event.''\}\\
\{``treatment'': ``What therapy was administered to the
patients?''\}\\
\{``subject-race'': ``What is the race or nationality of the
subject?''\}\\
\{``subject-disorder'': ``What preexisting conditions does the
subject have besides the main treatment disorder?''\}
\medskip

\textbf{Output JSON:}\\
\{\,``questions'': [``your set of questions'']\,\}
\end{promptbox}

\subsection{Question Refinement}
\begin{promptbox}
\label{prompt:question-refinement}
\small
You are a helpful assistant who refines questions to better extract
ground truth arguments for a given role. You previously generated a set of questions, but the extraction
results show room for improvement. Your task is to analyze the
feedback and produce an improved set of questions.
\medskip

\textbf{Context}\\
\textbf{Role:} \{role\}\\
\textbf{Document:} \{document\}\\
\textbf{Ground Truth Arguments:} \{gt\_arguments\}
\medskip

\textbf{Previous Questions (Iteration \{iteration\})}\\
\{current\_questions\}
\medskip

\textbf{Extraction Performance}\\
\textbf{Precision:} \{precision\} \quad
\textbf{Recall:} \{recall\} \quad
\textbf{F1-Score:} \{f1\_score\}
\medskip

\textbf{Summary:}\\
-- Successfully extracted: \{n\_matched\_gt\} out of \{n\_gt\}
ground truth arguments\\
-- Total predictions: \{n\_pred\} arguments\\
-- Correct predictions: \{n\_matched\_pred\} out of \{n\_pred\}
\medskip

\textbf{What Was Matched}\\
These arguments were successfully extracted:\\
\{matched\_arguments\}
\medskip

\textbf{What Was Missed}\\
These ground truth arguments were NOT extracted:\\
\{missing\_arguments\}
\medskip

\textbf{What Was Over-Generated}\\
These arguments were predicted but are NOT in the ground truth
(hallucinations or irrelevant extractions):\\
\{extra\_arguments\}
\medskip

\textbf{Your Task}\\
Based on the feedback above, \textbf{refine the question set} to:
\begin{enumerate}
\item \textbf{Add or modify questions} to capture the missing
arguments
\item \textbf{Refine or remove questions} that led to
over-generation or hallucinations
\item \textbf{Keep questions} that successfully extracted matched
arguments (unless they also caused over-generation)
\item \textbf{Ensure clarity and specificity}---vague questions
often lead to irrelevant extractions
\end{enumerate}
\medskip

\textbf{Analysis Strategy}
\medskip

\textit{For Missing Arguments:}\\
-- Ask: ``What specific question would directly prompt extraction
of this argument from the document?''\\
-- Consider: Is the information in the document? If yes, what
keywords or phrases would target it?\\
-- Action: Add a new targeted question OR modify an existing
question to be more comprehensive
\medskip

\textit{For Over-Generated Arguments:}\\
-- Ask: ``Which question caused this hallucination or irrelevant
extraction?''\\
-- Consider: Is the question too broad, ambiguous, or leading?\\
-- Action: Make the question more specific, add constraints, or
remove it if redundant
\medskip

\textit{For Matched Arguments:}\\
-- Ask: ``Which question successfully extracted this?''\\
-- Action: Keep it, but check if it also caused over-generation
\medskip

\textbf{Guidelines:}\\
-- Avoid yes/no questions\\
-- Keep questions concise and unambiguous\\
-- Do not create redundant questions---each question should have a
clear, distinct purpose\\
-- The total number of questions does not have to stay the same;
focus on \textbf{quality over quantity}
\medskip

\textbf{Critical Rules}
\medskip

\textbf{Absolutely forbidden} (violation makes questions useless
for generalization):\\
1.~Copying specific facts from ``What Was Missed'' into questions\\
2.~Turning missed information into yes/no questions\\
3.~Embedding document-specific details in questions\\
4.~Asking ``Did [specific fact] happen?''\\
5.~Leaking ground-truth arguments through generated questions\\
6.~Any other disguised leakage
\medskip

\textbf{Required:}\\
1.~Questions must target information \textit{categories}, not
specific values\\
2.~Questions must be generic enough to work on different documents\\
3.~Questions must be clear about what \textit{type} of information
is wanted
\medskip

\textbf{Output Format}\\
Return your response \textbf{strictly in JSON format} as:\\
\{``questions'': [``your refined set of questions''],
``reasoning'': ``Brief explanation of key changes and why''\}
\end{promptbox}

\subsection{Prediction}

\begin{promptbox}
\label{prompt:prediction}
\small
You are a helpful assistant who \textbf{concisely} extracts
arguments for a specified role from a given document, following the
role question.
\medskip

\textbf{Role to extract:} \{role\}\\
\textbf{Role-specific Questions:} \{role\_question\}\\
\textbf{Document:} \{document\}
\medskip

\textbf{Instructions:}\\
-- Extract only the arguments relevant to the given role
from the document following the role question.\\
-- If no argument is present, return ``null''.\\
-- If multiple arguments are present, separate them with a
semicolon (;).\\
-- Do not include explanations, reasoning, or any extra text.\\
-- Return the output strictly in JSON format.\\
-- Do not repeat any part of the question.\\
-- \textbf{The JSON must have exactly one key: the role name from
``Role to extract'' above. Do NOT use the word ``role'' as the
key.} Example: if Role to extract is ``Patient'', output
\{``Patient'': ``arg1; arg2''\}. If Role to extract is
``Age-at-Presentation'', output \{``Age-at-Presentation'':
``arg1; arg2''\}.
\medskip

\textbf{Output JSON} (use the role value from ``Role to extract''
as the key):\\
\{\,``\{role\}'': ``arg1; arg2''\,\}
\end{promptbox}

\section{Auto-Interpretability}
\label{section:interp}

Below we give examples of automatically interpreting how questions transform using the matryoshka methodology from \citet{cherep2026visualpersuasion}. We use an LLM-as-judge to label pairs of (Vanilla, Zero-Shot) or (Zero-Shot, Optimized) questions to interpret what changes, agglomeratively clustering embeddings of these differences and summarizing the clusters to the top-level summaries shown below. ``Lexical'' below means with token-level edit operations; otherwise diffs come from LLM-as-judge.

\setlength{\tabcolsep}{2pt}
\renewcommand{\arraystretch}{1}
\small
\setlist[itemize]{leftmargin=*, nosep, topsep=1pt, partopsep=0pt, parsep=0pt}
\onecolumn
\begin{longtable}{@{}p{0.1\textwidth}p{0.29\textwidth}p{0.29\textwidth}p{0.29\textwidth}@{}}
\toprule
Dataset & Vanilla $\to$ Zero-Shot (LLM) & Zero-Shot $\to$ Optimized (Lexical) & Zero-Shot $\to$ Optimized (LLM) \\
\midrule
\endfirsthead
\toprule
Dataset & Vanilla $\to$ Zero-Shot (LLM) & Zero-Shot $\to$ Optimized (Lexical) & Zero-Shot $\to$ Optimized (LLM) \\
\midrule
\endhead
CRB & \begin{itemize}
\item \textbf{Scope narrowing to clinical features} -- \textit{Focus shifts from broad symptoms/signs to precise neurological, anatomical, ocular, or vital}
\item \textbf{Addition of temporal and treatment context} -- \textit{Inclusion of timing, treatment, or situational clinical scenarios in questions}
\item \textbf{Scope narrowing to specific entities and terminology} -- \textit{From broad categories to particular therapies, disorders, tests; includes terminology generalization}
\item \textbf{Question format and response constraint changes} -- \textit{Shift from open-ended to yes/no formats and addition of temporal constraints}
\end{itemize} & \begin{itemize}
\item \textbf{Lexical substitution and refinement} -- \textit{Replacing words or phrases with more precise, formal, or clinically focused alternatives}
\item \textbf{Insertion of clarifying or specifying terms} -- \textit{Adding words to specify context, timing, case references, or details}
\item \textbf{Deletion of redundant or less relevant text} -- \textit{Removing unnecessary words or phrases to streamline questions}
\item \textbf{Grammatical and structural adjustments} -- \textit{Singular-plural form changes and question phrasing or temporal expression modifications}
\end{itemize} & \begin{itemize}
\item \textbf{Clinical scope expansion} -- \textit{Adds symptoms, diagnostics, family history, outcomes, interventions with risk}
\item \textbf{Anatomical and biochemical specification} -- \textit{Includes brain regions, metabolites, lab findings, detailed anatomical features}
\item \textbf{Terminology refinement and formalization} -- \textit{Refines clinical/genetic terms, clarifies inheritance, replaces informal terms}
\item \textbf{Question format and wording adjustments} -- \textit{Shifts query style, splits questions, adds adjectives, clarifies timing and sources}
\end{itemize} \\
\midrule
DEE & \begin{itemize}
\item \textbf{Subject perspective and terminology shifts} -- \textit{Changes between third-person and second-person subjects; 'subject' to 'patient' terminology}
\item \textbf{Temporal focus adjustments} -- \textit{Shifts from specific past/future events to broader or current time frames; additions}
\item \textbf{Scope narrowing and broadening} -- \textit{Changes between specific medication/substance and general drug categories or treatment status; shifts}
\item \textbf{Question format and focus shifts} -- \textit{Changing question types (yes/no to open-ended) and shifting emphasis among tapering, dosage}
\end{itemize} & \begin{itemize}
\item \textbf{Medication term generalization} -- \textit{Replacing specific drug names and dosages with general medication or treatment terms}
\item \textbf{Dosage and frequency phrasing changes} -- \textit{Modifying how dosage amounts and intake frequency are expressed or queried}
\item \textbf{Lexical substitution for question framing} -- \textit{Replacing interrogatives with descriptive or indirect wording and adding procedural details}
\item \textbf{Symptom and condition lexical refinement} -- \textit{Replacing general terms with specific symptom phrases and adjusting risk and temporal}
\end{itemize} & \begin{itemize}
\item \textbf{Scope broadening or narrowing} -- \textit{Changes in question focus between specific drugs/dosages and broader medication categories}
\item \textbf{Comparison target adjustment} -- \textit{Shifts in substances, conditions, or subjects compared or referenced}
\item \textbf{Temporal specification and perspective shifts} -- \textit{Introduction or refinement of timeframes and changes in temporal viewpoint}
\item \textbf{Supervision and tapering plan constraints} -- \textit{Includes lack of medical supervision and detailed tapering plan elements; medication type}
\end{itemize} \\
\midrule
MBAT & \begin{itemize}
\item \textbf{Scope narrowing to specific clinical details} -- \textit{From general tests, procedures, and symptoms to precise measurements, assays, imaging features,}
\item \textbf{Disease and symptom specificity} -- \textit{General disease terms replaced with specific diseases; symptoms refined with situational, temporal,}
\item \textbf{Medication detail specification} -- \textit{Specifies exact medication names or classes, dosage, regimen, treatment context, and temporal}
\item \textbf{Question format and context shifts} -- \textit{Addition of temporal constraints, patient context, and shift to closed yes/no or}
\end{itemize} & \begin{itemize}
\item \textbf{Lexical substitution for specificity} -- \textit{Replacing vague terms with precise clinical or descriptive words}
\item \textbf{Interrogative form modification} -- \textit{Changing question words, phrasing, structure, and punctuation including synonyms}
\item \textbf{Lexical insertion and deletion} -- \textit{Adding or removing descriptive, temporal, or passive terms}
\item \textbf{Question phrasing shift} -- \textit{Changing open-ended questions to closed or explicit formats, including modality changes}
\end{itemize} & \begin{itemize}
\item \textbf{Numeric to qualitative shift} -- \textit{From exact numeric values to qualitative or evaluative answers}
\item \textbf{Open-ended to closed-ended format} -- \textit{Changed question style from open response to yes/no format}
\item \textbf{Terminology refinement and generalization} -- \textit{Rephrasing specific terms to broader or more precise descriptors, including technical terms}
\item \textbf{Scope modification and narrowing} -- \textit{Changes in entity focus, timing, anatomy, symptom triggers; removal of patient references;}
\end{itemize} \\
\midrule
PHEE & \begin{itemize}
\item \textbf{Patient population specificity} -- \textit{General individual references changed to specific patient groups or conditions}
\item \textbf{Specified medical entities} -- \textit{Medication or condition explicitly named (e.g., ampicillin, retinopathy, meningitis)}
\item \textbf{Shift to interrogative questions} -- \textit{Changed from general statements or noun phrases to explicit question forms}
\item \textbf{Disorder category consolidation and wording simplification} -- \textit{Combining/reducing disorder categories; removing explanatory clauses; replacing 'subject' with 'patient(s)'; simplifying phrasing}
\end{itemize} & \begin{itemize}
\item \textbf{Lexical substitution and insertion} -- \textit{Replacing or adding terms to refine focus or increase specificity}
\item \textbf{Insertion of qualifiers and descriptors} -- \textit{Adding adjectives, numeric qualifiers, or clinical details to entities}
\item \textbf{Addition of clarifying phrases} -- \textit{Including explicit question components or clarifying clauses}
\item \textbf{Specification of subject and condition} -- \textit{Clarifying species, cases, or condition references within questions}
\end{itemize} & \begin{itemize}
\item \textbf{Patient group focus shifts} -- \textit{Changes in referenced patient populations and clinical attribute emphasis}
\item \textbf{Adverse effect and therapy specificity} -- \textit{Identification of specific adverse effects and explicit drug or agent naming}
\item \textbf{Causation versus association framing} -- \textit{Shift from general therapy association to specific causative agent inquiry}
\item \textbf{Scope modification and question clarity} -- \textit{Scope narrowing or broadening by disorder/treatment specificity and qualifier removal; question wording}
\end{itemize} \\
\bottomrule
\end{longtable}

\end{document}